\documentclass[journal]{IEEEtran}
\usepackage{graphicx}
\usepackage{epsfig}
\usepackage{epstopdf}
\usepackage{subfigure}
\usepackage{amsmath,amssymb} % define this before the line numbering.
\usepackage{color}
\usepackage{times}
\usepackage{xcolor}
\usepackage{subfigure}
\usepackage{rotating, multirow}
\usepackage{slashbox}
\usepackage{multirow}
\usepackage{multicol}
\usepackage{amsmath}
\usepackage{amssymb}
\usepackage{latexsym}
\usepackage{longtable}
\usepackage{color}
\usepackage{capt-of}
\usepackage{supertabular,booktabs}
\usepackage{enumerate}
\usepackage{enumitem}
\usepackage{enumitem}
\usepackage{makecell}
\setenumerate[1]{itemsep=0pt,partopsep=0pt,parsep=\parskip,topsep=5pt}
\setitemize[1]{iteFmsep=0pt,partopsep=0pt,parsep=\parskip,topsep=5pt}
\setdescription{itemsep=0pt,partopsep=0pt,parsep=\parskip,topsep=5pt}

\usepackage{soul}
\usepackage{enumitem}
\usepackage[marginal]{footmisc}
\usepackage{cite}
\usepackage[utf8]{inputenc}
\usepackage[small]{caption}
\usepackage{diagbox}
\usepackage{longtable}
\usepackage{supertabular}
\usepackage{float}
\usepackage{CJK}
\usepackage{url}
\usepackage[colorlinks,
            linkcolor=red,
            anchorcolor=blue,
            citecolor=green,
            backref=page
            ]{hyperref}
\usepackage{cleveref}
\begin{document}

%\title{A straw shows which way the wind blows: a survey for CNN-based density estimation and crowd counting}
\title{CNN-based Density Estimation and Crowd Counting: A Survey}

\author{Guangshuai Gao$^{1,2}$,
        Junyu Gao$^{3}$,~\IEEEmembership{Student Member,~IEEE},
        Qingjie Liu$^{1,2*}$,~\IEEEmembership{Member,~IEEE},
        Qi Wang$^{3}$,~\IEEEmembership{Senior Member,~IEEE},
        and Yunhong Wang$^{1,2}$,~\IEEEmembership{Fellow,~IEEE}

\thanks{Guangshuai Gao, Qingjie Liu and Yunhong Wang are with the State Key Laboratory of Virtual Reality Technology and Systems, Beihang University,
Xueyuan Road, Haidian District, Beijing, 100191, China and Hangzhou Innovation Institute, Beihang University, Hangzhou, 310051,China (email: gaoguangshuai1990@buaa.edu.cn; qingjie.liu@buaa.edu.cn;yhwang@buaa.edu.cn);

Junyu Gao and Qi Wang are with the School of Computer Science and with the Center for Optical Imagery Analysis and Learning (OPTIMAL), Northwestern Polytechnical University, Xi'an 710072, Shanxi, China (email: gjy3035@gmail.com;crabwq@gmail.com)

* Corresponding author: Qingjie Liu}}% <-this % stops a space

\maketitle

\begin{abstract}
Accurately estimating the number of objects in a single image is a challenging yet meaningful task and has been applied in many applications such as urban planning and public safety. In the various object counting tasks, crowd counting is particularly prominent due to its specific significance to social security and development. Fortunately, the development of the techniques for crowd counting can be generalized to other related fields such as vehicle counting and environment survey, if without taking their characteristics into account. Therefore, many researchers are devoting to crowd counting, and many excellent works of literature and works have spurted out. In these works, they are must be helpful for the development of crowd counting. However, the question we should consider is why they are effective for this task. Limited by the cost of time and energy, we cannot analyze all the algorithms. In this paper, we have surveyed over 220 works to comprehensively and systematically study the crowd counting models, mainly CNN-based density map estimation methods. Finally, according to the evaluation metrics, we select the top three performers on their crowd counting datasets and analyze their merits and drawbacks. Through our analysis, we expect to make reasonable inference and prediction for the future development of crowd counting, and meanwhile, it can also provide feasible solutions for the problem of object counting in other fields. We provide the density maps and prediction results of some mainstream algorithm in the validation set of NWPU dataset for comparison and testing. Meanwhile, density map generation and evaluation tools are also provided. All the codes and evaluation results are made publicly available at \url{https://github.com/gaoguangshuai/survey-for-crowd-counting}.
\end{abstract}

\begin{IEEEkeywords}
Object counting, crowd counting, density estimation, CNNs.
\end{IEEEkeywords}

\IEEEpeerreviewmaketitle

\section{Introduction}
\IEEEPARstart{O}{ver} the past few decades, an increasing number of research communities, have considered the problem of object counting as their mainly research direction, as a consequence, many works have been published to count the number of objects in images or videos across wide variety of domains such as crowding counting~\cite{zhang2016single,onoro2016towards,boominathan2016crowdnet,kang2018crowd,sam2017switching,sindagi2017generating,liu2018decidenet,hossain2019crowd,zhang2018crowd,sang2019improved,cao2018scale,li2018csrnet,varior2019scale}, cell microscopy~\cite{wang2016fast,walach2016learning,lempitsky2010learning}, animals~\cite{arteta2016counting}, vehicles~\cite{onoro2016towards,zhang2017visual,zhang2017fcn,guerrero2015extremely}, leaves~\cite{aich2017leaf,giuffrida2016learning} and environment survey~\cite{french2015convolutional,zhan2008crowd}. In all these domains, crowd counting is of paramount importance, and it is crucial to building a more high-level cognitive ability in some crowd scenarios, such as crowd analysis~\cite{shao2015deeply,zhou2012understanding} and video surveillance~\cite{chan2008privacy}. As the increasing growth of the world's population and subsequent urbanization result in a rapid crowd gathering in many scenarios such as parades, concerts and stadiums. In these scenarios, crowd counting plays an indispensable role for social safety and control management.

Considering the specific importance of crowd counting aforementioned, more and more researchers have attempted to design various sophisticated projects to address the problem of crowd counting. Especially in the last half decades, with the advent of deep learning, Convolution Neural Networks (CNNs) based models have been overwhelmingly dominated in various computer vision tasks, including crowd counting. Although different tasks have their unique attributes, there exist common features such as structural features and distribution patterns. Fortunately, the techniques for crowd counting can be extended to some other fields with specific tools. Therefore, in this paper, we expect to provide a reasonable solution for other tasks through the deep excavation of the crowd counting task, especially for CNN-based density estimation and crowd counting models. Our survey aims to involve various parts, which is ranging algorithm taxonomy from some interesting under-explored research direction. Beyond taxonomically reviewing existing CNN-based crowd counting and density estimation models, representing datasets and evaluation metrics, some factors and attributes, which largely affect the performance the designed model, are also investigated, such as distractors and negative samples. We provide the density maps and prediction results of some mainstream algorithm in the validation set of NWPU dataset~\cite{wang2020nwpu} for comparison and testing. Meanwhile, density map generation and evaluation tools are also provided. All the codes and evaluation results are made publicly available at \url{https://github.com/gaoguangshuai/survey-for-crowd-counting}.

\subsection{Related Works and Scope}
The various approaches for crowd counting are mainly divided into four categories: detection-based, regression-based, density estimation, and more recently CNN-based density estimation approaches. We focus on the CNN-based density estimation and crowd counting model in this survey. For the sake of completeness, it is necessary to review some other related works in this subsection.

Early works~\cite{topkaya2014counting,li2008estimating,leibe2005pedestrian,enzweiler2009monocular} on crowd counting use detection-based approaches. These approaches usually apply a person or head detector via a sliding window on an image. Recently many extraordinary object detectors such as R-CNN~\cite{girshick2015fast,ren2015faster,he2017mask},  YOLO~\cite{redmon2016you}, and SSD~\cite{liu2016ssd} have been presented, which may perform dramatic detection accuracy in the sparse scenes. However, they will present unsatisfactory results when encountered the situation of occlusion and background clutter in extremely dense crowds.

To reduce the above problems, some works~\cite{idrees2013multi,chan2008privacy,chan2009bayesian} introduce regression-based methods which directly learn the mapping from an image patch to the count. They usually first extract global features~\cite{chen2012feature} (texture, gradient, edge features), or local features~\cite{ryan2009crowd} (SIFT~\cite{lowe1999object}, LBP~\cite{ojala2000gray}, HOG~\cite{dalal2005histograms}, GLCM~\cite{haralick1973textural}). Then some regression techniques such as linear regression~\cite{paragios2001mrf} and Gaussian mixture regression~\cite{tian2010latent} are used to learn a mapping function to the crowd counting.

These methods are successful in dealing with the problems of occlusion and background clutter, but they always ignore spatial information. Therefore, Lemptisky et al.~\cite{lempitsky2010learning} first adopt a density estimation based method by learning a linear mapping between local features and corresponding density maps. For reducing the difficulty of learning a linear mapping, \cite{pham2015count} proposes a non-linear mapping, random forest regression, which obtains satisfactory performance by introducing a crowdedness prior and using it to train two different forests. Besides, this method needs less memory to store the forest. These methods consider the spatial information, but they only use traditional hand-crafted features to extract low-level information, which cannot guide the high-quality density map to estimate more accurate counting.

\begin{figure*}[t]
%%tr = 0.006, ts = 0.008
  \centering
      \centerline{\includegraphics[width=0.99 \linewidth]{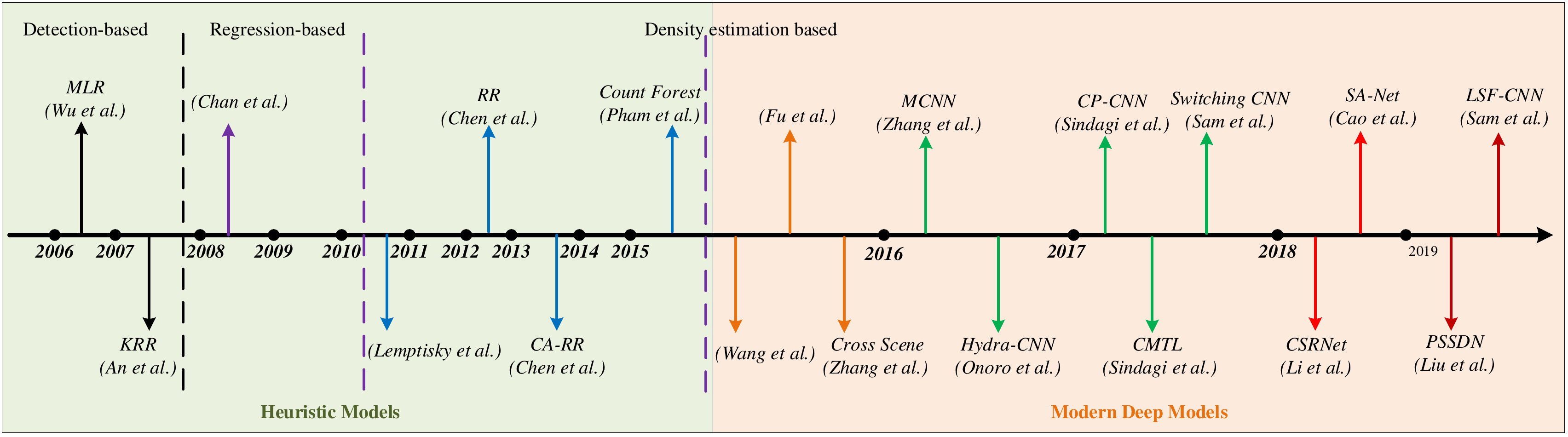}}
\caption{A brief chronology of crowd counting. The first incorporation of deep learning techniques for crowd counting is from 2015. See Section 1 for more detailed description. Milestone models in this figure: MLR~\cite{wu2006crowd}, KRR~\cite{an2007face}, Chan et al.~\cite{chan2008privacy}, Lemptisky et al.~\cite{lempitsky2010learning}, RR~\cite{chen2012feature}, CA-RR~\cite{chen2013cumulative}, Count Forest~\cite{pham2015count}, Wang et al.~\cite{wang2015deep}, Fu et al.~\cite{fu2015fast}, Cross scene~\cite{zhang2015cross}, MCNN~\cite{zhang2016single}, Hydra-CNN~\cite{onoro2016towards}, CP-CNN~\cite{sindagi2017generating}, CMTL~\cite{sindagi2017cnn}, switching CNN~\cite{sam2017switching}, CSRNet~\cite{li2018csrnet}, SANet~\cite{cao2018scale}, PSSDN~\cite{liu2019point} and LSF-CNN~\cite{sam2019locate}. The trend in the past few years has been designing crowd counting models based on multi-column (in green), single-column (in red) network architecture and object localization or tracking depending on counting techniques (in crimson), which are either contemporary and potential direction in future. While traditional heuristic methods are highlighted with the blue-shaded area and the modern CNN-based density estimation and crowd counting models are with the red-shaded backgrounds, respectively.}
\vspace{-5pt}
\label{fig:chronology}
\end{figure*}

\begin{figure*}[t]
%%tr = 0.006, ts = 0.008
  \centering
      \centerline{\includegraphics[width=1.0 \linewidth]{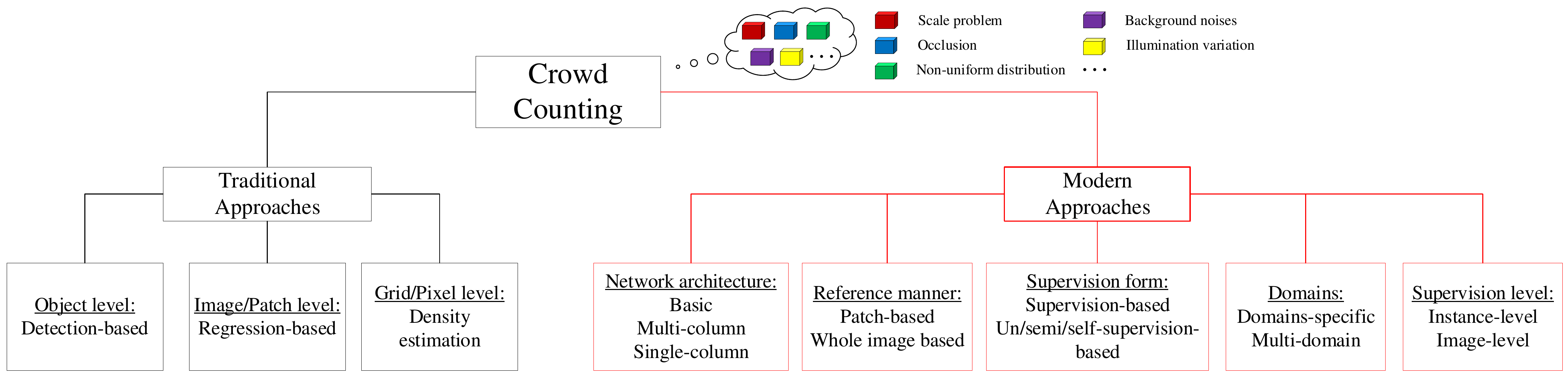}}
\caption{The overall architecture of this work. We concentrate on the modern density map-based approaches mainly CNN-based for crowd counting.}
\vspace{-5pt}
\label{fig:pipline}
\end{figure*}

Recently, benefiting from the powerful feature representation of CNNs, more researchers utilize it to improve the density estimation. Earlier heuristic models typically leverage basic CNNs to predict the density of the crowds~\cite{wang2015deep,fu2015fast,zhang2015cross,walach2016learning}, which obtain significant improvement compared with traditional hand-crafted features. Lately, more effective and efficient models based on Fully Convolution Network (FCN), which has become the mainstream network architecture for the density estimation and crowd counting. Different supervised level and learning paradigm for different models, also there are some models designed in cross scene and multiple domains. A brief chronology is shown in Fig.~\ref{fig:chronology}, which illustrates the main advancements and milestones of crowd counting techniques. The goal of this survey is focused on the modern CNN-based for density estimation and crowd counting, Fig.~\ref{fig:pipline} depicts a taxonomy of curial methodologies to be covered in the survey.

\textbf{Scope of the survey.} Considering that reviewing all state-of-the-art methods is impractical (and fortunately unnecessary), this paper sorts out some mainstream algorithms, which are all influential or essential papers published in, but not limited to, prestigious journals and conferences. The survey focuses on the modern CNN-based density estimation methods in recent years, and some early works are also included for the sake of completeness. We classify existing methods into several categories, in terms of network architecture, supervision form, influence of cross-scene or multi-domain, etc. Such comprehensive and systematic taxonomies can be more helpful for the readers to in-depth understand the progress of crowd counting in the past years.

\begin{figure*}[t]
%%tr = 0.006, ts = 0.008
  \centering
      \centerline{\includegraphics[width=1.0 \linewidth]{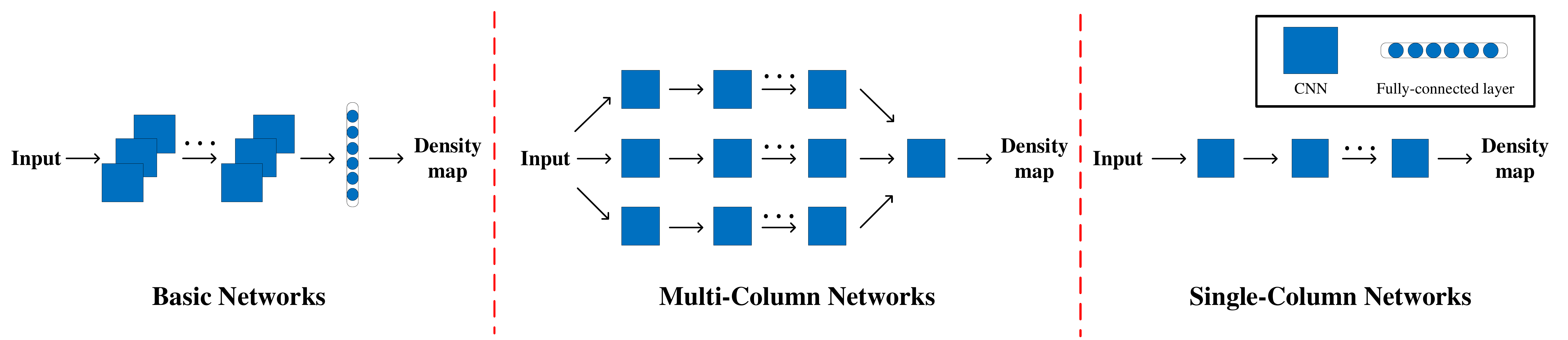}}
\caption{Comparison of the structure of existing density map-based networks.}
\vspace{-5pt}
\label{fig:network}
\end{figure*}

\subsection{Related previous reviews and surveys}

\begin{table*}[!htb]
    \scriptsize
	\caption{Summary of previous reviews.}
	\vspace{-0.3cm}
	\begin{center}
		\renewcommand{\arraystretch}{1.5}	
		\setlength\tabcolsep{5pt}
    \begin{tabular}{|l|p{5cm}|l|l|p{8cm}|}
    \hline
    \# & Title & Year & Venue & Brief description \\  \hline
    1 & Crowd analysis: a survey~\cite{zhan2008crowd} & 2008 & MVA &This paper presents a survey on crowd analysis methods employed in computer vision research and discusses perspectives from other research disciplines and how they can contribute to the computer vision approach. \\ \hline
    2 & Crowd analysis using computer vision techniques~\cite{junior2010crowd} & 2010 & ISPM & A survey on crowd analysis by using computer vision techniques, including different aspects such as people tracking, crowd density estimation, event detection, validation and simulation. \\ \hline
    3 &A Survey of Human-Sensing:Methods for Detecting Presence, Count, Location, Track, and Identity~\cite{teixeira2010survey} &2010 &ACM Computing Surveys&a survey of the inherently multidisciplinary literature of human-sensing , focusing mainly on the extraction of five commonly needed spatio-temporal properties: namely presence, count, location, track and identity. \\ \hline
    4 & Crowd counting and profiling: Methodology and evaluation~\cite{loy2013crowd} & 2013 & MSVAC & This study describes and compares the state-of-the-art methods for video imagery based crowd counting, and provides a systematic evaluation of different methods using the same protocol. \\ \hline
    5 & Performance evaluation of crowd image analysis using the PETS2009 dataset~\cite{ferryman2014performance} & 2014 & PRL & This paper presents PETS2009 crowd analysis dataset and highlights detection and tracking performance on it\\ \hline
    6 & Crowded scene analysis: A survey~\cite{li2015crowded} & 2015 & TCSVT & This paper surveys the state-of-the-art techniques on crowded scene analysis with different methods such as crowd motion pattern learning, crowd behavior, activity analysis and anomaly detection in crowds. \\ \hline
    7 & An evaluation of crowd counting methods, features and regression models~\cite{ryan2015evaluation} & 2015 & CVIU & This paper presents an evaluation across multiple datasets to compare holistic, local and histogram based methods, and to compare various image features and regression models. \\ \hline
    8 & Recent survey on crowd density estimation and counting for visual surveillance~\cite{saleh2015recent} & 2015 & EAAI & This paper presents a survey on crowd density estimation and counting methods employed for visual surveillance in the perspective of computer vision research. \\ \hline
    9 & Advances and trends in visual crowd analysis: A systematic survey and evaluation of crowd modelling techniques~\cite{zitouni2016advances} & 2016 & Neurocomputing & This paper aims to give an account of such issues by deducing key statistical evidence from the existing literature and providing recommendations towards focusing on the general aspects of techniques rather than any specific algorithm. \\ \hline
    10 &Crowd scene understanding from video: a survey~\cite{grant2017crowd} & 2017 & TOMM & This survey explores crowd analysis as it relates to two primary research areas: crowd statistics and behavior understanding.  \\ \hline
    11 & A survey of recent advances in cnn-based single image crowd counting and density estimation~\cite{sindagi2018survey} & 2018 & PRL & A review of various single image crowd counting and density estimation methods with a specific focus on recent CNN-based approaches. \\ \hline
    %12 & Beyond Counting: Comparisons of Density Maps for Crowd Analysis Tasks - Counting, Detection, and Tracking~\cite{kang2018beyond} & 2018 & TCSVT & Comparison of density map estimation methods and their performance on counting and two localization tasks, detection and tracking.\\
    12 &Convolutional neural networks for crowd behaviour analysis: a survey~\cite{tripathi2019convolutional} &2019 &VC &A survey for crowd analysis using CNN \\ \hline
    %\hline
    \end{tabular}
    \end{center}
    \label{survey}
\end{table*}

Table~\ref{survey} lists the existing reviews or surveys which are related to our paper. Notably, Zhan et al~\cite{zhan2008crowd} and Junior et al.~\cite{junior2010crowd} are the first ones for crowd analysis. Li et al.~\cite{li2015crowded} review the task of crowded scene analysis with different methods, while Zitouni et al.~\cite{zitouni2016advances} evaluate different methods with different criteria. Loy et al.~\cite{loy2013crowd} make detailed comparisons of state-of-the-arts for crowd counting based on video imagery with the same protocol. Ryan et al.~\cite{loy2013crowd} present an evaluation across multiple datasets to compare various image features and regression models and Saleh et al.~\cite{saleh2015recent} survey two main approaches in direct and indirect manners. Grant et al.~\cite{grant2017crowd} explore two kinds of crowd analysis. While these surveys make detail analysis on crowd counting and scene analysis, they are only for traditional methods with hand-crafted features. In recent work, Sindagi et al.~\cite{sindagi2018survey} provide a survey of recent state-of-the-art CNN-based approaches for crowd counting and density estimation for the single image. However, it only roughly introduces the latest advancement of CNN-based methods, which are only up to the year 2017. Tripathi et al.~\cite{tripathi2019convolutional} put forward a review on crowd analysis using CNN, which is not just for crowd counting, thereby it was not adequate comprehensive and in-depth. As we know, the techniques are incremental month by month, and it is also an urgent need for us to document the development of crowd counting in the past half-decade.

Different from previous surveys that focus on hand-crafted features or primitive CNNs, our work systematically and comprehensively reviews CNN-based density estimation crowd counting approaches. Specifically, we summarize the existing crowd counting models from various aspects and list the results of some representing mainstream algorithms in terms of evaluation metrics on several typical benchmark crowd counting datasets. Finally, we select the top three performers and carefully and thoroughly analyze the properties of these models. We also offer insights for essential open issues, challenges, and future direction. Through this survey, we expect to make reasonable inference and prediction for the future development of crowd counting, and meanwhile, it can also provide feasible solutions and make guidance for the problem of object counting in other domains.

\subsection{Contributions of this paper}

In summary, the contributions in this paper are mainly in the following folds:

\begin{enumerate}
\item \textbf{Comprehensive and systematic overview from various aspects}. We category the CNN-based models according to several taxonomies, including network architecture, supervised form, learning paradigm, etc. The taxonomies can motivate researches with a deep understanding of the critical techniques of CNN-based methods.

\item \textbf{Attribute-based performance analysis}. Based on the performance of the SOTA methods,  we analyze the reasons why they perform well, the techniques they utilize. Besides, we discuss the various challenge factors that promote researchers to design more effective algorithms.

\item \textbf{Open questions and future directions.} We look through some important issues for model design, dataset collection, and some generalization to other domains with domain adaptation or transfer learning and explore some promising research directions in the future.
\end{enumerate}

These contributions provide detailed and in-depth review, which differs from the previous review or survey works to a large extent.

The remainder of the paper is organized as follows. Section~\ref{section:mainstream} conducts a comprehensive literature review of mainstream CNN-based density estimation and crowd counting models according to the proposed taxonomies. Section~\ref{section:datasets} examines the most notable datasets for crowd counting and some datasets for other object counting tasks, while section~\ref{section:metrics} describes several widely used evaluation metrics. Section~\ref{section:benchmarking} benchmarks some representing models and makes an in-depth analysis. Section~\ref{section:discussion} presents a discussion and put forward some open issues and possible future directions. Finally, the conclusion is concluded in Section~\ref{section:conclusion}.

%---------------------------------------------------------------------------------------------------------------------------------------------------------------------------------------------------------------------------------%
\begin{table*}[!htb]
\scriptsize
\caption{Summary of state-of-the-art methods. See~\ref{section:mainstream} for more detailed description.}
\begin{center}
\begin{tabular}{|r|l|c|c|c|c|c|}
\hline
Methods &Year$\&$Venue &Network architecture &Reference manner &Supervision form & Learning paradigm &Supervision level  \\ \hline \hline
Fu et al.~\cite{fu2015fast} &2015 EAAI &Basic & Patch-based &Fully-Sup. &STL &Instance level \\ \hline
Wang et al.~\cite{wang2015deep} &2015 ACMMM &Basic &Patch-based &Fully-Sup. &STL &Instance level \\ \hline
Cross scene~\cite{zhang2015cross} &2015 CVPR &Basic &Patch-based &Fully-Sup. &MTL &Instance level \\ \hline
\hline
MCNN~\cite{zhang2016single} &2016 CVPR &Multi-column &Whole image-based &Fully-Sup. &STL &Instance level  \\ \hline
Crowdnet~\cite{boominathan2016crowdnet} &2016 ACMMM &Multi-column &Patch-based &Fully-Sup. &STL &Instance level \\ \hline
CNN-Boosting~\cite{walach2016learning} &2016 ECCV &Basic &Patch-based &Fully-Sup. &STL &Instance level \\ \hline
Hydra-CNN~\cite{onoro2016towards} &2016 ECCV &Multi-column &Patch-based &Fully-Sup. &MTL &Instance level \\ \hline
Shang et al.~\cite{shang2016end} &2016 ECCV &Multi-column &Whole image-based &Fully-Sup. &STL &Instance level \\ \hline
\hline
%MoC-CNN~\cite{kumagai2017mixture} &2017 arxiv &Multi-column &Patch-based &Fully-Sup. &STL &Instance level \\ \hline
%FCNCC~\cite{marsden2017fully} &2017 VISAPP &Single-column &Whole image-based &Fully-Sup. &STL &Instance level \\ \hline
CMTL~\cite{sindagi2017cnn} &2017 AVSS &Multi-column &Whole image-based &Fully-Sup. &MTL &Instance level \\ \hline
Switching CNN~\cite{sam2017switching} &2017 CVPR &Multi-column &Patch-based &Fully-Sup. &MTL &Instance level \\ \hline
%MSCNN~\cite{zeng2017multi} &2017 ICIP &Single-column &Whole image-based &Fully-Sup. &STL &Instance level \\ \hline
CP-CNN~\cite{sindagi2017generating} &2017 ICCV &Multi-column &Whole image-based &Fully-Sup. &MTL &Instance level \\ \hline
\hline
%AMDCN~\cite{deb2018aggregated} &2018 CVPRW &Multi-column &Whole image-based &Fully-Sup. &STL &Instance level \\ \hline
D-ConvNet~\cite{shi2018crowd} &2018 CVPR &Single-column &Whole image-based &Filly-Sup. &STL &Instance level \\ \hline
CSRNet~\cite{li2018csrnet}&2018 CVPR &Single-column &Whole image-based &Fully-Sup. &STL &Instance level \\ \hline
DRSAN~\cite{liu2018crowd} &2018 IJCAI &Multi-column &Whole image-based &Fully-Sup. &STL &Instance level \\\hline
DecideNet~\cite{liu2018decidenet} &2018 CVPR &Multi-column &Patch-based &Fully-Sup. &MTL &Instance level \\ \hline
%SCNet~\cite{wang2018defense} &2018 BMVC &Single-column &Whole image-based &Fully-Sup. &STL &Instance level \\ \hline
SaCNN~\cite{zhang2018crowd} &2018 WACV &Single column &Whole image-based &Fully-Sup. &MTL &Instance level \\ \hline
SACNN~\cite{cao2018scale} &2018 ECCV &Single column &Patch-based &Fully-Sup. &MTL &Instance level \\ \hline
IG-CNN~\cite{babu2018divide} &2018 CVPR &Multi-column &Patch-based &Fully-Sup. &MTL &Instance level \\ \hline
ic-CNN~\cite{ranjan2018iterative} &2018 ECCV &Multi-column &Whole image-based &Fully-Sup. &MTL &Instance level \\ \hline
ACSCP~\cite{shen2018crowd} &2018 CVPR &Multi-column &Patch-based &Fully-Sup. &MTL &Instance level \\ \hline
NetVLAD~\cite{shi2018multiscale} &2018 TII &Single-column &Whole image-based &Fully-Sup. &MTL &Instance level \\ \hline
CL~\cite{idrees2018composition} &2018 ECCV &Single-column &Patch-based &Fully-Sup. &MTL &Instance level \\ \hline
L2R~\cite{liu2018leveraging} &2018 CVPR &Basic &Whole image-based &Self-Sup. &MTL &-- \\ \hline
GAN-MTR~\cite{olmschenk2018crowd} &2018 WACV &Basic &Whole image-based &Semi-Sup. &MTL &-- \\ \hline
%CCLL~\cite{zhou2018crowd} &2018 T-ITS &Non-deep &Patch-based &Semi-Sup. &STL &-- \\ \hline
\hline
%SDA-MCNN~\cite{yang2019counting} &2019 Neurocomputing &Multi-column &Patch-based &Fully-Sup. &MTL &Instance level \\ \hline
PaDNet~\cite{tian2019padnet} &2019 TIP &Single-column &Patch-based &Fully-Sup. &STL &Instance level \\ \hline
ASD~\cite{wu2019adaptive} &2019 ICASSP &Multi-column &Whole image-based &Fully-Sup. &MTL &Instance level \\ \hline
SPN~\cite{chen2019scale} &2019 WACV &Single column &Whole image-based &Fully-Sup. &STL &Instance level \\ \hline
SR-GAN~\cite{olmschenk2019generalizing} &2019 CVIU &Basic &Whole image-based &Semi-Sup. &MTL &-- \\ \hline
ADCrowdnet~\cite{liu2019adcrowdnet} &2019 CVPR &Single column &Whole image-based &Fully-Sup. &STL &Instance level \\ \hline
SAAN~\cite{hossain2019crowd} &2019 WACV &Multi-column &Whole image-based &Fully-Sup. &MTL &Instance level \\ \hline
SAA-Net~\cite{varior2019scale} &2019 CVPR &Single column &Whole image-based &Fully-Sup. &MTL &Instance level \\ \hline
SFCN$\dag^2$~\cite{wang2019learning} &2019 CVPR &Single column &Whole image-based &Fully-Sup. &STL &Instance level \\ \hline
SE Cycle GAN~\cite{wang2019learning} &2019 CVPR &Single column &Whole image-based &Fully-Sup. &STL &Instance level \\ \hline
PACNN~\cite{shi2019revisiting} &2019 CVPR &Single column &Whole image-based &Fully-Sup. &STL &Instance level \\ \hline
CAN\&ECAN~\cite{liu2019context} &2019 CVPR &Single column &Whole image-based &Fully-Sup. &STL &Instance level \\ \hline
%DENet~\cite{liu2019denet} &2019 arxiv &Multi-column &Whole image-based &Fully-Sup. &MTL &Instance level \\ \hline
CFF~\cite{shi2019counting} &2019 ICCV &Single-column &Whole image-based &Fully-Sup. &MTL &Instance level \\ \hline
%DSNet~\cite{dai2019dense} &2019 arxiv &Single-column &Whole image-based &Fully-Sup. &STL &Instance level  \\ \hline
PCC Net~\cite{gao2019perspective} &2019 TCSVT &Multi-column &Whole image-based &Fully-Sup. &MTL &Instance level \\ \hline
SFANet~\cite{zhu2019dual} &2019 CVPR &Single column &Whole image-based  &Fully-Sup. &MTL &Instance level \\ \hline
W-Net~\cite{valloli2019w} &2019 CVPR &Single column &Whole image-based  &Fully-Sup. &STL &Instance level \\ \hline
SL2R~\cite{liu2019exploiting} &2019 CVPR &Basic &Whole image-based &Self-Sup. &MTL &-- \\ \hline
%DG-GAN~\cite{olmschenk2019dense} &2019 CVPRW &Basic &Whole image-based &Semi-Sup. &MTL &-- \\ \hline
TEDnet~\cite{jiang2019crowd} &2019 CVPR &Single column &Whole image-based &Fully-Sup. &STL &Instance level \\ \hline
RReg~\cite{wan2019residual} &2019 CVPR &Multi-column &Whole image-based &Fully-Sup. &STL &Instance level \\ \hline
RAZNet~\cite{liu2019recurrent} &2019 CVPR &Multi-column &Whole image-based &Fully-Sup. &MTL &Instance level \\ \hline
AT-CNN~\cite{zhao2019leveraging} &2019 CVPR &Single-column &Whole image-based &Fully-Sup. &MTL &Instance level \\ \hline
GWTA-CCNN~\cite{deepak2019almost} &2019 AAAI &Single column &Patch-based &Un-Sup. &STL &-- \\ \hline
HA-CCN~\cite{sindagi2019ha-ccn} &2019 TIP &Single column &Whole image-based &Fully-Sup./Weak-Sup &STL &Instance/Image level \\ \hline
L2SM~\cite{xu2019learn} &2019 ICCV &Single column &Patch-based &Fully-Sup &STL &Instance level \\ \hline
RANet~\cite{zhang2019relational} &2019 ICCV &Multi-column &Whole image-based &Fully-Sup &STL &Instance level \\ \hline
McML~\cite{cheng2019improving} &2019 ACM MM &Multi-column &Whole image-based &Fully-Sup &STL &Instance level \\ \hline
%MTCNet~\cite{kumar2019mtcnet} &2019 AVSS &Single column &Whole image-based &Fully-Sup &MTL &Instance level \\ \hline
ILC~\cite{cholakkal2019object} &2019 CVPR &Multi-column &Whole image-based  &Fully-Sup. &MTL &Image level \\ \hline
\end{tabular}
\end{center}
\label{table:summary}
\end{table*}

\section{Taxonomy for crowd counting}
\label{section:mainstream}
In this section, we review CNN-based crowd counting algorithms in the following taxonomies. Chiefly is representative network architectures for crowd counting (\ref{representative}). Next is the learning paradigm of the methods (\ref{sub:paradigm}), and then is the inference manner of the networks (\ref{inference}). Additionally, the supervision forms of networks are also introduced in \ref{supervision}. Meanwhile, to evaluate the generalization ability of the algorithms, we classify existing works into domain-specific and multi-domain ones (\ref{domain}). Finally, based on the supervised level, we classify the CNN-based models into instance-level and image-level ones (\ref{instance}). We group the important models and describe them roughly in chronological order. A summary of the state-of-the-art is presented in Table~\ref{table:summary}.
\subsection{Representative network architectures for crowd counting}
\label{representative}
In view of different types of network architectures, we divide crowd counting models into three categories: basic CNN based methods, multi-column based methods, and single-column based methods. The category of network architectures is illustrated in Fig.~\ref{fig:network}.

%\begin{enumerate}[leftmargin=*]
\subsubsection{\textbf{Basic CNN}}
%\item \textbf{Basic CNN network architectures}
\label{sub:basic}

This network architecture adopts the basic CNN layers which convolutional layers, pooling layers, uniquely fully connected layers, without additional feature information required. They generally are involved in the initial works using CNN for density estimation and crowd counting.

\noindent $\bullet$ \textbf{Fu et al.}~\cite{fu2015fast} put forward the first CNN-based model for crowd counting, which accelerates the speed and accuracy of the model by removing some similar network connections existed in feature maps and cascading two ConvNet classifiers.

\noindent $\bullet$ \textbf{Wang et al.}~\cite{wang2015deep} propose a deep network based on Alexnet architecture~\cite{krizhevsky2012imagenet} for extremely dense crowd counting, the adoption of expanded negative samples, whose ground truth counting are zeros, to reduce the interference.

\noindent $\bullet$ \textbf{CNN-boosting}~\cite{walach2016learning} employs basic CNNs in a layer-wise manner, and leverages layered boosting and selective sampling to improve the counting accuracy and reduce training time.

Since without additional feature information provided, basic CNNs are simple and easy to implement yet usually perform low accuracy.

\subsubsection{\textbf{Multi-column}}
\label{sub:multi-column}

These network architectures usually adopt different columns to capture multi-scale information corresponding to different receptive fields, which have brought about excellent performance for crowd counting.

\noindent $\bullet$ \textbf{MCNN}~\cite{zhang2016single}, a pioneering work explicitly focusing on the multi-scale problem. MCNN is a multi-column architecture with three branches that use different kernel sizes (large, medium, small). However, the similar even the same depth and structure of the three branches, which makes the network look like a simple assembling of several weak regressors.

\noindent $\bullet$ \textbf{Hydra-CNN}~\cite{onoro2016towards} uses a pyramid of image patches corresponding to different scales to learn a multi-scale non-linear regression model for the final density map estimation.

\noindent $\bullet$ \textbf{CrowdNet}~\cite{boominathan2016crowdnet} combines shallow and deep networks at different columns, of which the shallow one captures the low-level features corresponding to large scale variation and the deep one captures the high-level semantic information.

%\noindent $\bullet$ \textbf{MoC-CNN}~\cite{kumagai2017mixture} combines multiple expert CNNs to address the appearance variations, each of which is specialized to a specific appearance. For the adaptively selecting the expert CNNs, a gating CNN is incorporated to assign different weights on them.

\noindent $\bullet$ \textbf{Switching CNN}~\cite{sam2017switching} trains several independent CNN crowd density regressors on the image patches, the regressors have the same structure with MCNN~\cite{zhang2016single}. In addition, a switch classifier is also trained alternatively on the regressions to select the best one for the density estimation.

\noindent $\bullet$ \textbf{CP-CNN}~\cite{sindagi2017generating} is a contextual pyramid CNN that combines global and local contextual information to generate high-quality density maps. Moreover, adversarial learning ~\cite{goodfellow2014generative} is utilized to fuse the features from different levels.

%\noindent $\bullet$ \textbf{AFS-FCN}~\cite{kang2018crowd} leverages an image pyramid manner to carry out each image with different sizes into a Fully Convolutional Network (FCN) to predict density maps. Besides, an attention map is weighted on each prediction map to model scale changes at different locations of the images.

\noindent $\bullet$ \textbf{TDF-CNN}~\cite{sam2018top} delivers top-down information to the bottom-up network to amend the density estimation.

%\noindent $\bullet$ \textbf{AMDCN}~\cite{deb2018aggregated} assembles the multi-scale information generated by the multi-column CNNs to increase the performance.

\noindent $\bullet$ \textbf{DRSAN}~\cite{liu2018crowd} handles the issues of scale variation and rotation variation taking advantages of Spatial Transformer Network (STN)~\cite{jaderberg2015spatial}.

\noindent $\bullet$ \textbf{SAAN}~\cite{hossain2019crowd} is similar to the idea of MoC-CNN~\cite{kumagai2017mixture} and CP-CNN~\cite{sindagi2017generating}, but utilizes visual attention mechanism to automatically select the particular scale both for the global image level and local image patch level.

\noindent $\bullet$ \textbf{RANet}~\cite{zhang2019relational} provides local self-attention (LSA) and global self-attention (GSA) to capture short-range and long-range interdependence information respectively, furthermore, a relation module is introduced to merge LSA and GSA to obtain more informative aggregated feature representations.

\noindent $\bullet$ \textbf{McML}~\cite{cheng2019improving} incorporates a statistical network into the multi-column network to estimate the mutual information between different columns, the proposed mutual learning scheme which can optimize each column alternately whilst retaining other columns fixed on each mini-batch training data.

\noindent $\bullet$ \textbf{DADNet}~\cite{guo2019dadnet} takes dilated-CNN with different dilated rates to capture more contextual information as front-end and adaptive deformable convolution as a back-end to locate the positions of the objects accurately.

%\noindent $\bullet$ \textbf{SDA-MCNN}~\cite{yang2019counting} adopts similar structure as MCNN~\cite{zhang2016single} to capture multi-scale and a weighted Euclidean loss to handle non-uniform crowd distributions. Additionally, perspective maps in dense crowds are estimated in accord with head-head distance.

%\noindent $\bullet$ \textbf{DeepCount}~\cite{chen2019deep} regresses a global count trained from density map supervision by introducing multi-layer gradient fusion strategy.

%\noindent $\bullet$ \textbf{ACM-CNN}~\cite{zou2019attend} puts forward the first attempt to introduce the adaptive capacity model for crowd counting, and novel count attention is adopted to select crowd regions without priors.

Albeit great progress has been achieved by these multi-column network, they still suffer from several significant disadvantages, which have been demonstrated through conducting experiments by Li et al.~\cite{li2018csrnet}. First of all, it is difficult to train the multi-column networks since it requires more time and a more bloated structure. Next, using different branches but almost the same network structures, it inevitably leads to a lot of information redundancy. Moreover, multi-column networks always require density-level classifiers before sending images into the networks. However, due to the number of crowds is varying greatly in the congested scene of the real world, making it difficult to define the granularity of density level. Meanwhile, more fine-grained classifiers also mean that more columns and more sophisticated structures are required to be designed, thereby causing more redundancy. Finally, these networks consume a large number of parameters for density-level classifiers rather than preparing them for the generation of final density maps. Thus the lack of parameters for density map generation will degrade the quality.

As all the disadvantages mentioned above, multi-column network architectures may be ineffective in a narrow sense. Thus it motivates many researchers to exploit simpler yet effective and efficient networks. Therefore, single column network architectures are come out to cater to the demands of more challenging situations in the crowd counting.
%%%------------------------------------------------------------------------------------------------------------------------------------------------------------------------------%%%

\subsubsection{\textbf{Single column}}
\label{sub:single}

The single-column network architectures usually deploy single and deeper CNNs rather than the bloated structure of multi-column network architecture, and the premise is not to increase the complexity of the network.

%\noindent $\bullet$ \textbf{FCNCC}~\cite{marsden2017fully} is a deep, single column, fully convolutional network to estimate the density map, during inference a multi-scale averaging step is used to address the scale and perspective distortion problem.

%\noindent $\bullet$ \textbf{MSCNN}~\cite{zeng2017multi} incorporates a multi-scale blob with different kernel sizes which is similar to the network structure of naive Inception module~\cite{szegedy2015going}.

\noindent $\bullet$ \textbf{W-VLAD}~\cite{sheng2018crowd} takes account of semantic features and spatial cues, additionally, a novel locality-aware feature (LAF) is introduced to represent the spatial information.

\noindent $\bullet$ \textbf{SaCNN}~\cite{zhang2018crowd} is a scale-adaptive CNN that takes an FCN with fixed small receptive fields as backbone and adapts the feature maps extracted from multiple layers to the same sizes and then combines them to generate the final density map.

%\noindent $\bullet$ \textbf{SCNet}~\cite{wang2018defense} cascades three modules named as residual fusion module, pyramid pooling module, and sub-pixel convolutional module into a unified single-column network.

\noindent $\bullet$ \textbf{D-ConvNet}~\cite{shi2018crowd} called as De-correlated ConvNet, takes advantage of negative correlation learning (NCL) to improve the generalization capability of the ensemble models with a set of weak regressors with convolutional feature maps.

\noindent $\bullet$ \textbf{CSRNet}~\cite{li2018csrnet} adopts dilated convolution layers to expand the receptive field while maintaining the resolution as back-end network.

\noindent $\bullet$ \textbf{SANet}~\cite{cao2018scale} is built on the shoulder of Inception architecture~\cite{szegedy2015going} in the encoder to extract multi-scale features and using Transposed convolution layers in the decoder to up-sampling the extracted feature maps.

%\noindent $\bullet$ \textbf{Improved SaCNN}~\cite{sang2019improved} is an improved method based on SaCNN~\cite{zhang2018crowd} via optimizing the standard variance of geometry-adaptive Gaussian kernel~\cite{zhang2016single} to obtain precise head size and high-quality ground truth density map.

%\noindent $\bullet$ \textbf{DA-Net}~\cite{zou2018net} first adopts deformable convolutions in the crowd counting task to improve the robustness for scale-variation of the model. Different connections are adopted compared with SaCNN~\cite{zhang2018crowd} and adaptive weights are assigned to different layers according to their importance.

\noindent $\bullet$ \textbf{SPN}~\cite{chen2019scale} leverages a shared deep single-column structure and extracts the multi-scale features in the high-layers by Scale Pyramid Module (SPM), which deploys four parallel dilated convolution with different dilation rates.

\noindent $\bullet$ \textbf{ADCrowdNet}~\cite{liu2019adcrowdnet} combines visual attention mechanism and multi-scale deformable convolutional scheme into a cascading framework.

\noindent $\bullet$ \textbf{SAA-Net}~\cite{varior2019scale} mimics multi-branches but single column by learning a set of soft gate attention mask on the intermediate feature maps, which uses the hierarchical structure of CNNs. The ides behind it is somewhat similar to SaCNN~\cite{zhang2018crowd} but adding attention mask on corresponding feature maps.

\noindent $\bullet$ \textbf{W-Net}~\cite{valloli2019w} is inspired by U-Net~\cite{ronneberger2015u}, adding an auxiliary Reinforcement branch to accelerate the convergence and retain local pattern consistency, and using Structural Similarity Index (SSIM) to estimate the final density maps. %Considering that in the decoding up-sampling phase in the SANet~\cite{cao2018scale}, the attended transposed layers may bring checkerboard artifacts into the mix making it a soup sandwich, W-Net uses an operation of nearest-neighbor interpolation as a better alternative, as suggested in ~\cite{odena2016deconvolution}.

\noindent $\bullet$ \textbf{TEDnet}~\cite{jiang2019crowd} is a trellis encoder-decoder network architecture, which integrates multiple decoding paths to capture multi-scale features and exploits dense skip connections to obtain the supervised information. In addition, to alleviate the gradient vanishing problem and improve the back-propagation ability, a combinational loss comprising local coherence and spatial correlation loss is also presented.

%\noindent $\bullet$ \textbf{DSNet}~\cite{dai2019dense} proposes a novel dense dilated convolution block (DDCB), which cascades dilated convolution layers with different dilated rates and dense residual connection. Moreover, a similar multi-scale density level consistency loss is also introduced to boost performance.

%\noindent $\bullet$ \textbf{DDCN}~\cite{wang2019removing} aims to reduce the effect of background interference from the detail layer of the network. Moreover, a weighted Euclidean loss is devised to allocate different weights on the crowd and the background separately, which further boosts the counting performance.

Due to their architectural simplicity and training efficiency, single column network architecture has received more and more attention in the recent years.

\subsection{Learning paradigm}
\label{sub:paradigm}
From the view of different paradigms, crowd counting networks can be bifurcated as single-task and multi-task based methods.

\subsubsection{\textbf{Single-task based methods}}

The classical methodology is to learn one task at one time, i.e., single-task learning~\cite{caruana1997multitask}. Most CNN-based crowd counting methods belong to this paradigm, which generally generates density maps and then sum all the pixels to obtain the total count number, or the count number directly.

\subsubsection{\textbf{Multi-task based methods}}

More recently, inspired by the success of multi-task learning in various computer vision tasks, it has shown better performance by combing density estimation and other tasks such as classification, detection, segmentation, etc. Multi-task based methods are generally designed with multiple subnets; besides, in contrast to pure single column architecture, there may be other branches corresponding to different tasks. In summary, multi-task architectures can be regarded as the cross-fertilize between multi-column and single-column but different from either one.

\noindent $\bullet$ \textbf{CMTL}~\cite{sindagi2017cnn} combines crowd count classification and density map estimation into an end-to-end cascaded framework. It divides crowd count into groups and takes this as a high-level prior to integrate into the density map estimation network.

\noindent $\bullet$ \textbf{Decidenet}~\cite{liu2018decidenet} predicts the crowd count by generating the detection-based and regression-based density maps, respectively. To adaptively decide which model is appropriate, an attention module is adopted to guide the network to allocate relative weights and further select suitable mode. It can automatically switch between detection and regression mode. However, it may suffer from a huge number of parameters by utilizing the multi-column structure.

\noindent $\bullet$ \textbf{IG-CNN}~\cite{babu2018divide} is a hierarchical clustering model, which can generate image groups in the dataset and a set of particular networks specialized in their respective group. It can adapt and grow regarding the complexity of the dataset.

\noindent $\bullet$ \textbf{ic-CNN}~\cite{ranjan2018iterative} puts forward a two-branch network, one of which is generating low-resolution density maps, and the other is refining the low-resolution maps and feature maps extracted from previous layers to produce higher resolution density maps.

\noindent $\bullet$ \textbf{ACSCP}~\cite{shen2018crowd} ACSCP introduces an adversarial loss to make the blurring density maps sharp. Moreover, a scale-consistency regularizer is designed to guarantee the calibration of cross-scale model and collaboration between different scale paths.

\noindent $\bullet$ \textbf{CL}~\cite{idrees2018composition} simultaneously addresses three tasks, including crowd counting, density map estimation, and localization in dense crowds, according to the fact that they are related to each other making the loss function in the optimization of deep CNN decomposable.

%\noindent $\bullet$ \textbf{ASD}~\cite{wu2019adaptive} designs a model of scenario discovery and modeling crowd counting simultaneously, which is composed of two parallel pathways to serve multi-scale feature extraction and crowd densities, both of them are convolution layers with different sizes of receptive fields. Scenario can be considered as the combination of two pathways with some discrete responses which are learned from an adaptive branch.

%\noindent $\bullet$ \textbf{SFANet}~\cite{zhu2019dual} attempts to combines visual attention mechanism and multi-scale feature fusion into a dual path framework, one is for the attention map generation and the other is for fusing multi-scale features and then incorporating attention map to generate the final density map.

%\noindent $\bullet$ \textbf{DENet}~\cite{liu2019denet} first takes Mask R-CNN~\cite{he2017mask} as detector to segment the objects which are clearly distinct from crowd, and after removing the segmented areas, uses a modified Xception~\cite{isola2017image} as the encoder to extract multi-scale features and combines dilated convolution and transposed convolution for the generation of density maps. Compared with the computation cost of DecideNet~\cite{liu2018decidenet}, DENet is simpler yet effective and efficient.

\noindent $\bullet$ \textbf{CFF}~\cite{shi2019counting} assumes that point annotations not just for constructing density maps, repurposing the point annotations for free in two ways. One is supervised focus from segmentation, and the other is from global density. The focus for free can be regarded as the complement of other excellent approaches, which benefits counting if ignoring the base network.

\noindent $\bullet$ \textbf{PCC Net}~\cite{gao2019perspective} takes perspective change into account, which is composed of three components, Density Map Estimation for leaning local features, Random High-level Density Classification for predicting density labels of image patches, and Fore-/Background Segmentation (FBS) for segmenting the foreground and background.

\noindent $\bullet$ \textbf{RAZ-Net}~\cite{liu2019recurrent} observes that the density map is not consistent with the correct person density, which implies that crowd localization cannot depend on the density map. A recurrent attentive zooming network is proposed to increase the resolution for localization and an adaptive fusion strategy to enhance the mutual ability between counting and localization.

\noindent $\bullet$ \textbf{ATCNN}~\cite{zhao2019leveraging} fuses three heterogenous attributes, i.e., geometric, semantic and numeric attributes, taking them as auxiliary tasks to assist the crowd counting task.

\noindent $\bullet$ \textbf{CDT}~\cite{kang2018beyond} not only makes an overall comparison of density maps on counting, but also extends to detection and tracking.

%\noindent $\bullet$ \textbf{MTCNet}~\cite{kumar2019mtcnet} tackles two related tasks including Crowd Density Estimation and Crowd-Count Group Classification, simultaneously.

%\noindent $\bullet$ \textbf{MRA-CNN}~\cite{zhang2019multi} deals with the counting task and additional density-level classification task, multi-scale and multi-contextual features are extracted to tackle the problem of scale variation and non-uniform distribution, a multi-resolution attention model is also exploited to cope with the intricate backgrounds.

\noindent $\bullet$ \textbf{NetVLAD}~\cite{shi2018multiscale,zhang2019nonlinear} is a multi-scale and multi-task framework which assembles multi-scale features captured from the input image into a compact feature vector in the means of "Vector of Locally Aggregated Descriptors" (VLAD). Additionally, "deeply supervised" operations are exploited on the bottom layers to provide additional information to boost the performance.

%\noindent $\bullet$ \textbf{DNCL}~\cite{zhang2019nonlinear} is an extension journal version of ~\cite{shi2018crowd}, which extends the original work to deal with more regression based tasks, including crowd counting, age estimation, personality analysis and image super-resolution.

%\noindent $\bullet$ \textbf{DaD}~\cite{liu2019using} trains a two-stream network to regress both the people density and the scene depth, which is effective for resist adversarial attacks in the deep crowd counting models.

%\noindent $\bullet$ \textbf{STANet}~\cite{wen2019drone} is a space-time multi-scale attention network which aggregates multiple feature maps in successive frames by utilizing the temporal consistency, then jointly addresses the issues of density map prediction, object localization and object tracking in drone-based images with arbitrary density, perspective, and flight altitude.

%\noindent $\bullet$ \textbf{CCCNet}~\cite{das2019cccnet} opens up a new variant of crowd counting called as categorized crowd counting, which provides an attention-based network to tackle crowd counting meanwhile pose estimation such as whether sitting or standing of persons.

\subsection{Inference manner}
\label{inference}
Based on the different training manners, the CNN-based crowd counting approaches can be classified as patch-based inference and the whole image-based inference.

\subsubsection{\textbf{Patch-based methods}}

This inference manner is required to train using patches randomly cropped from the image. In the test phase, using a sliding window spreads over the whole test image, and getting the estimations of each window and then assembling them to obtain the final total count of the image.

\noindent $\bullet$ \textbf{Cross-scene}~\cite{zhang2015cross} randomly selects overlapping patches from the training images to serve as training samples, and the density maps of corresponding image patches are treated as the ground truth. The total count of the selected training patch is computed by integrating over the density map. The value of count is a decimal, rather than an integer.

\noindent $\bullet$ \textbf{CCNN}~\cite{onoro2016towards} is primarily leaning a regression function to project the appearance of the image patches onto their corresponding object density maps. The model adopts the same sizes of all patches and the same covariance value of the Gaussian function in the groundtruth density map generation process, which limits the accuracy when encounters the large scale variation scenarios.

%\noindent $\bullet$ \textbf{A-CCNN}~\cite{amirgholipour2018ccnn} makes an improvement in CCNN~\cite{onoro2016towards} by using pathes with variable sizes and different covariance values to adapt various training patches.

\noindent $\bullet$ \textbf{DML}~\cite{wang2018deep} integrates metric learning into a deep regression network, which can simultaneously extract density-level features and learn better distance measurement.

\noindent $\bullet$ \textbf{PaDNet}~\cite{tian2019padnet} present a novel Density-Aware Network (DAN) module to discriminate variable density of the crowds, and Feature Enhancement Layer (FEL) module is to boost the global and local recognition performance. %In addition, Patch MAE and Patch MSE are used to evaluate both global and local accuracy and robustness.

\noindent $\bullet$ \textbf{L2SM}~\cite{xu2019learn,xu2019autoscale} attempts to address the density pattern shift issue, which is resulting from nonuniform density between sparse and dense regions, by providing two modules, i.e., Scale Sreserving Network (SPN) to obtain patch-level density maps and a learn to scale module (L2SM) to compute scale ratios for dense regions. %In addition, a multipolar center loss (MPCL) is adopted to normalize the density maps automatically. %AutoScale~\cite{xu2019autoscale} extends L2SM into a common FPN-like framework, which can select the dense regions adaptively. Moreover, a novel distance label map combine with an adapted cross-entropy loss are introduced to facilitate the accurate person localization. Additionally, it generalizes the model to vehicle counting.

\noindent $\bullet$ \textbf{GSP}~\cite{aich2019global} devises a global sum pooling operation to replace global average pooling (GAP) or fully connected layers (FC), considering the counting task as a simple linear mapping problem and avoiding patchwise cancellation and overfitting in the training phase with small datasets of large images.

\subsubsection{\textbf{Whole image-based methods}}

Patch-based methods always neglect global information and burden much computation cost due to the sliding window operation. Thus the whole image-based methods usually take the whole image as input, and output corresponding density map or a total number of the crowds, which is more convergence but may lose local information sometimes.

\noindent $\bullet$ \textbf{JLLG}~\cite{shang2016end} feeds the whole image into a pre-trained CNN to obtain high-level features, then maps these features to local counting numbers. It takes advantage of contextual information both in the global and local count.

\noindent $\bullet$ \textbf{Weighted VLAD}~\cite{sheng2016crowd} integrates semantic information into learning locality-aware feature (LAF) sets for crowd counting. First, mapping the original pixel space onto a dense attribute feature map, then utilizing the LAF to capture more spatial context and local information.

\subsection{Supervision form}
\label{supervision}
According to whether human-labeled annotations are used for training, crowd counting methods can be classified into two categories: fully-supervised methods and un-/self-/semi-supervised methods.

\subsubsection{\textbf{Fully-supervised methods}}

The vast majority of CNN-based crowd counting methods rely on large-scale accurately hand-annotated and diversified data. However, the acquisition of these data is a time-consuming and more onerous labeling burden than usual. Beyond that, due to the rarely labeled data, the methods may suffer from the problem of over-fitting, which leads to a significant degradation in performance when transferring them in the wild or other domains. Therefore, training data with less or even without labeled annotations is a promising research topic in the future.

\subsubsection{\textbf{Un/semi/weakly/self-supervised methods}}

Un/semi-supervised learning denotes that learning without or with a few ground-truth labels, while self-supervised learning represents that adding an auxiliary task which is different from but related to supervised tasks. Some methods exploit unlabeled data for training have achieved comparative performance in contrast with supervised methods.

\noindent $\bullet$ \textbf{GWTA-CCNN}~\cite{deepak2019almost} presents a stacked convolution autoencoder based on Grid Winner-Take-All~\cite{makhzani2015winner} paradigm for unsupervised feature learning, of which 99\% parameters can be trained without any labeled data.

\noindent $\bullet$ \textbf{SR-GAN}~\cite{olmschenk2019generalizing} generalizes semi-supervised GANs from classification problems to regression problems by introducing a loss function of feature contrasting.

\noindent $\bullet$ \textbf{GAN-MTR}~\cite{olmschenk2018crowd} applies semi-supervised learning GANs objectives to multiple object regression problem, which trains a basic network the same as \cite{zhang2015cross} with the use of unlabeled data.

\noindent $\bullet$ \textbf{DG-GAN}~\cite{olmschenk2019dense} presents a semi-supervised dual-goal GAN framework to seek both the number of individuals in the crowd scene and discriminate whether the real or fake images.

\noindent $\bullet$ \textbf{CCLL}~\cite{zhou2018crowd} puts forward a semi-supervised method by utilizing a sub-modular to choose the most representative frames from the sequences to circumvent redundancy and retain densities, graph Laplacian regularization and spatiotemporal constraints are also incorporated into the model.

\noindent $\bullet$ \textbf{L2R}~\cite{liu2018leveraging,liu2019exploiting} exploits unlabeled crowd data for pre-training CNNs in a multi-task framework, which is inspired by self-supervised learning and based on the observation that the crowd count number of the patches must be fewer or equal to the larger patch which contains them. The method is fully supervised in essence but an additional task of count ranking in a self-supervised manner. %SL2R~\cite{liu2019exploiting} extends the work by presenting a general framework for training from self-supervised ranking, which can be treated as a measure for active learning, expecting to achieve automatically choosing which image required to be labeled from abundant unlabeled data, so that eases the labeling burden.

\noindent $\bullet$ \textbf{HA-CNN}~\cite{sindagi2019ha-ccn} offers the first attempt to fine-turn the network to new scenes in a weakly supervised manner, by leveraging the image-level labels of crowd images into density levels.

\noindent $\bullet$ \textbf{CCWld}~\cite{wang2019learning} provides a data collector and labeler for crowd counting, where the data is from an electronic game. With the collector and labeler, it can collect and annotate data automatically, and the first large-scale synthetic crowd counting dataset is constructed.

\noindent $\bullet$ \textbf{CODA}~\cite{wang2019coda} presents a novel scale-aware adversarial density adaption approach for object counting, which can be used to generalize the trained model to unseen scenes in an unsupervised manner.

\noindent $\bullet$ \textbf{OSSS}~\cite{hossainone} designs a one-shot scene-specific crowd counting model by taking advantage of fine-turning.

%%---------------------------------------------------------------------------------------------------------------------------------------------------------------------------------%%%
\subsection{Domain adaptation}
\label{domain}
Almost all the existing counting methods are designed in a specific domain; therefore, designing a counting model which can count any object domain is a challenging yet meaningful task. The domain adaptation technique may be a powerful tool to tackle this problem.

\noindent $\bullet$ \textbf{CAC}~\cite{lu2018class} formulates the counting as a matching problem, which presents a Generic Matching Network (GMN) in a class-agnostic manner. GMN can be trained by the amount of video data labeled for tracking due to counting as a matching problem. In a few-shot learning way, it can use an adapter module to apply to different domains.

\noindent $\bullet$ \textbf{PPPD}~\cite{marsden2018people} provides a patch-based, multi-domain object counting network by leveraging a set of domain-specific scaling and normalization layers which only uses a few of parameters. It can also be extended to perform a visual domain classification even in an unseen observed domain.%, which outstands out its versatility and modular nature. The method has been successfully applied to people, penguins and cell counting.

\noindent $\bullet$ \textbf{SE CycleGAN}~\cite{wang2019learning} takes advantage of domain adaptation technique, incorporating Structural Similarity Index (SSIM)~\cite{wang2004image} into traditional CycleGAN framework to make up the domain gap between synthetic data and real-world data.

\noindent $\bullet$ \textbf{MFA+SDA}~\cite{gao2019feature} is drawing the idea from SE Cycle GAN, which is also a GAN-based adaptation model. The authors propose a Multi-level Feature-aware Adaptation to reduce the domain gap and present a Structured Density map Alignment for handling the unseen crowd scenes.

\noindent $\bullet$ \textbf{DACC}~\cite{gao2019domain} is composed of two modules: Inter-domain Features Segregation (IFS) and Gaussian-prior Reconstruction (GPR). IFS is designed to translate the synthetic data to realistic images, and GPR is used to generate higher-fidelity density maps with pseudo labels.

\noindent $\bullet$ \textbf{FSC}~\cite{han2020focus} extracts semantic domain-invariant features via crowd masks generated by a pre-trained crowd segmentation model. The error estimations in the background regions are reduced significantly.

\subsection{Instance-/image-based supervision}
\label{instance}
The aim of object counting is to estimate the number of objects. If the ground truth is labeled with point or bounding box, the method pertains to instance-level supervision. In contrast, image-level supervision just needs to count the number of different object instance instead.

\subsubsection{\textbf{Instance-level supervision}}

Most crowd density estimation methods are based on instance-level (point-level or bounding box) supervision, which needs hand-labeled annotations for each instance location.

\subsubsection{\textbf{Image-level supervision}}

Image-level supervision-based methods need to count the number of instances within or beyond the subitizing range, which do not require location information. It can be regarded as estimating the count at one shot or glance~\cite{chattopadhyay2017counting}.

\noindent $\bullet$ \textbf{ILC}~\cite{cholakkal2019object} generates a density map of object categories, which obtains the total object count estimation and spatial distribution of object instances simultaneously.

\section{Datasets}
\label{section:datasets}
\begin{figure*}[t]
%%tr = 0.006, ts = 0.008
  \centering
      \centerline{\includegraphics[width=1.0 \linewidth]{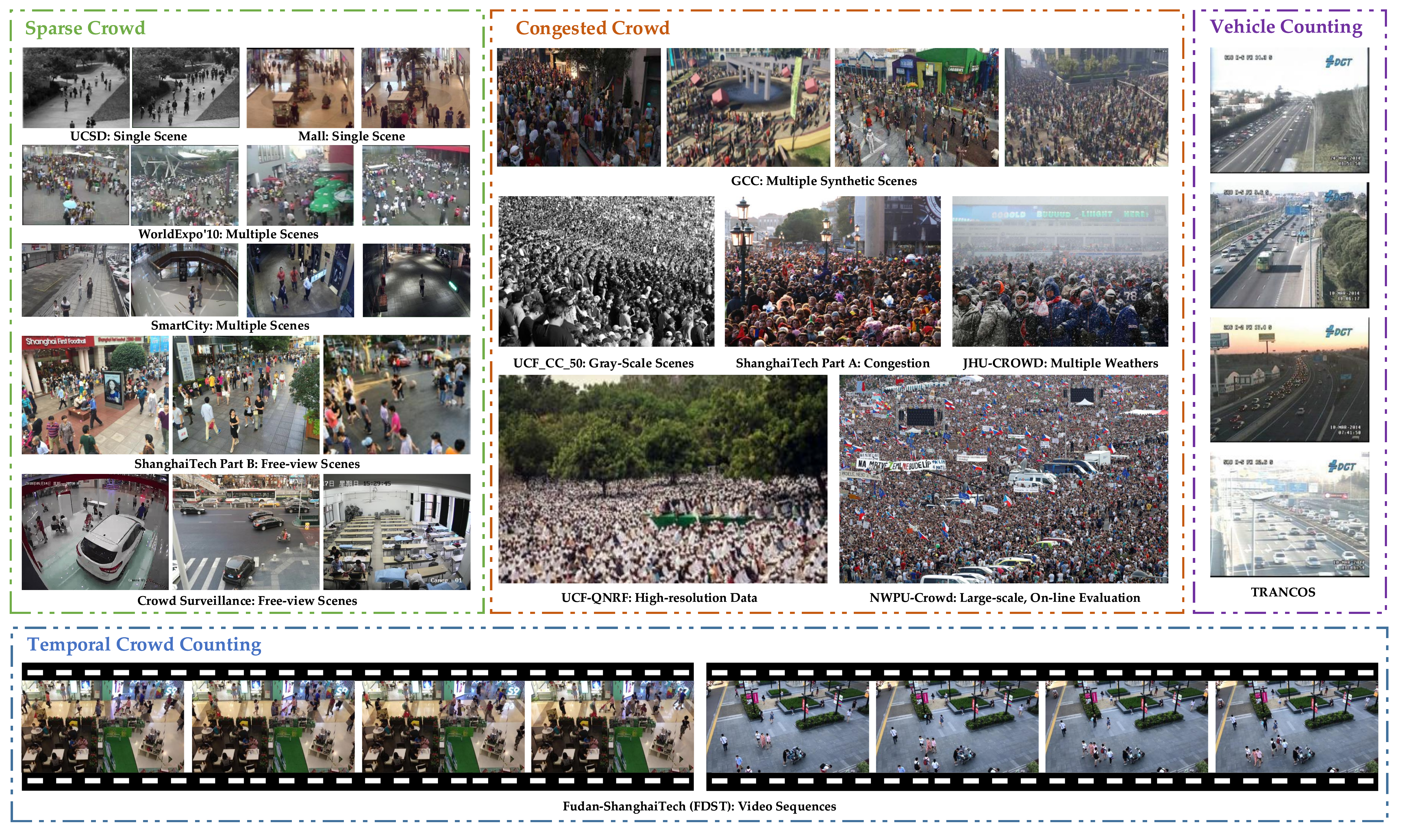}}
\caption{Some samples from representing crowd counting datasets.}
\vspace{-5pt}
\label{fig:datasets}
\end{figure*}

With the blooming development of crowd counting, numerous datasets have been introduced, which can motivate many more algorithms to cater to various challenges such as scale variations, background clutter in the surveillance video and changeable environment, illumination variation in the wild. In this section, we review almost all the crowd counting datasets from beginning up to now. Table~\ref{tabel:datasets} summarizes some representing datasets, including crowd counting datasets with real-world data and one with synthetic data, for the sake of completeness, we also survey several datasets applied in other domains, to evaluate the generalization ability of the designed algorithms. The datasets are sorted by chronology and the specific statistics of them are listed in Table~\ref{tabel:datasets}. Some samples from the representing datasets are depicted in Fig.~\ref{fig:datasets}.

\subsection{Most frequently-used datasets}
\label{most}
In this subsection, we introduce some most frequently used crowd counting datasets, i.e., UCSD~\cite{chan2008privacy}, Mall~\cite{chen2012feature}, UCF\_CC\_50~\cite{idrees2013multi}, WorldExpo'10~\cite{zhang2015cross}, Shanghai Tech~\cite{zhang2016single}, which are listed by chronologically.

\noindent $\bullet$ \textbf{UCSD~\cite{chan2008privacy}}\footnote{http://www.svcl.ucsd.edu/projects/anomaly/dataset.htm} is the first dataset for crowd counting, which is collected from cameras on the sidewalk. It is composed of 2000 frames with a size of 238$\times$158 and the ground truth annotations of each pedestrian in every five frames. For the rest of frames, the labels are created by using linear interpolation. %It also provides regions of interest to ignore the unnecessary moving objects such as tree. The dataset contains 49,885 pedestrians in total, which is divided into two parts: where 800 images which include 600 to 1399 pedestrians served as training set, while the remaining 1200 frames are served as test set. The dataset is low density with an average of 15 persons per frame.
Since it is collected from a single location, thus there is no change of the perspective view in different frames.

\noindent $\bullet$ \textbf{Mall~\cite{chen2012feature}}\footnote{http://personal.ie.cuhk.edu.hk/~ccloy/downloads\_mall\_dataset.html} is a dataset collected from the surveillance video of a shopping mall. The video sequence in the dataset is composed of 2000 frames with a size of 320$\times$240, which contains 62,325 pedestrians in total. %and the minimum, average and maximum number of people are 13, 31 and 53, respectively, and uses the first 800 frames as training set, remaining 1200 frames served as testing set.
Compared with UCSD~\cite{chan2008privacy}, Mall covers more diversity densities as well as different activity patterns (static and moving persons) under more significant illumination conditions. Additionally, there exists more perspective distortion, resulting in larger size change and appearance of objects, and has severe occlusions due to scene objects. %However, the dataset is a part of a single continuous video sequence, there is little change in the perspective view in different frames.

\noindent $\bullet$ \textbf{UCF\_CC\_50~\cite{idrees2013multi}}\footnote{https://github.com/davideverona/deep-crowd-counting\_crowdnet} is the first really challenging dataset created from publicly available Web images. It includes a variety of densities and different perspective distortions for different scenes such as concerts, protests, stadiums and marathons. %It contains a total of 50 images with different resolutions, with an average of 1,280 people per image. There are 63,075 people are annotated and the number of individuals is ranging from 94 to 4,543.
Considering that only 50 images in this dataset, a 5-fold cross-validation protocol is conducted on it. Due to the small-scale data volume, even the most advanced recent CNN-based methods are far from optimal for the results on it.

\noindent $\bullet$ \textbf{WorldExpo'10~\cite{zhang2015cross}}\footnote{http://www.ee.cuhk.edu.hk/~xgwang/expo.html} is a large data-driven cross-scene crowd counting dataset collected from Shanghai 2010 WorldExpo, which includes 1,132 annotated video sequences captured by 108 surveillance cameras. It contains a total of 3920 frames with a size of 576$\times$720, of which 199,923 persons are annotated. %The dataset is divided two parts: 1127 one-minute video sequences from 103 scenes were served as training set, whilst five one-hour video sequences from five different scenes. Each scene contains 120 labeled frames with a range from 1 to 220. Although attempting to capture different scenes from various density level, the test set only coveres five scenes with a maximum of 220 people, thus insufficient to evaluate high congested crowds.

\noindent $\bullet$ \textbf{Shanghai Tech~\cite{zhang2016single}}\footnote{https://pan.baidu.com/s/1nuAYslz} is one of the largest large-scale crowd counting datasets in previous few years which is composed of 1198 images with 330,165 annotations. According to different density distributions, the dataset is divided into two parts: Part\_A (SHT\_A) and Part\_B (SHT\_B). SHT\_A contains images randomly selected from the Internet, whilst Part\_B includes the images are taken from a busy street of a metropolitan area in Shanghai. The density in Part\_A is much larger than that in Part\_B. %These two parts are divided into training set and test set. Part\_A has 300 training images and 182 test images, while Part\_B has 400 training images and 316 test images.
This dataset successfully creates a challenging dataset across different scenes types and densities. However, the number of images in different density sets is uneven, which makes the training set and test set tend to be low-density sets. Nevertheless, the scale changes and perspective distortion presented by this dataset provide new challenges and opportunities for the design of many CNN-based networks.

\subsection{More recently datasets}
\label{more}

\noindent $\bullet$ \textbf{Smartcity~\cite{zhang2018crowd}}\footnote{https://pan.baidu.com/s/1pMuGyNp$\#$list/path=$\%$2F} is created by Tencent YouTu, which contains 50 images in 10 scenes such as sidewalk, office entrance, shopping mall. All of them are high shot for video surveillance. The dataset includes indoor and outdoor scenes, and mainly to verify the generalization ability of the model on very sparse scenes.

\noindent $\bullet$ \textbf{UCF-QNRF}~\cite{idrees2018composition}\footnote{https://www.crcv.ucf.edu/data/ucf-qnrf/} is collected from Flickr, Web Search and Hajj footage, which consists of 1,535 challenging images with about 1.25 million annotations. %Moreover, the average resolution of the images is 2013 $\times$ 2902, causing the absolute size of a person's head to vary drastically from a few pixels to more than 1500.
The images in this dataset come with a wider variety of scenes and contain the most diverse set of viewpoints, densities, and lighting variations. However, some of them are so high-resolution that they may lead to memory issues in GPU while training the entire scene.

\noindent $\bullet$ \textbf{City Street~\cite{zhang2019wide}} is a multi-view video dataset of which the data is collected from a busy city street by using five synchronized cameras, which is composed of 500 multi-view images in total %of which the first 300 are used for training and the rest 200 for testing.

\noindent $\bullet$ \textbf{ShanghaiTechRGBD~\cite{lian2019density}} is a large-scale RGB-D dataset which consists of 2,193 images with 144,512 labeled head counts. With the crowd scenarios and various lighting condition, making the dataset is the most challenging RGB-D crowd counting dataset regarding the number of head counts. %1,193 images are random selected for training and the remaining for testing.

\noindent $\bullet$ \textbf{FDST~\cite{fang2019locality}}\footnote{https://github.com/sweetyy83/Lstn\_fdst\_dataset} is a new large-scale video crowd counting dataset, which consists of 100 videos captured from 13 different scenes including shopping malls, squares, hospitals, etc, which contains 15,000 frames with 394,081 annotated heads, and all with frame-wise annotation. %In particular, 60 videos, 9000 frames are served as training set and the remain 40 videos, 6000 frames are served as testing set.

\noindent $\bullet$ \textbf{Crowd Surveillance~\cite{yan2019perspective}}\footnote{https://ai.baidu.com/broad/subordinate?dataset=crowd surv} contains 13,945 high-resolution images with 386,513 people in total, making the largest and highest average resolution for crowd counting so far. In addition, regions of interest (ROI) annotation is also provided to filter out the regions which are too blurry or ambiguous for training and test.

\noindent $\bullet$ \textbf{JHU-CROWD~\cite{sindagi2019pushing}} is a larger dataset w.r.t the number of images and several particular properties such as adverse conditions (weather-based degradations), learning bias mitigated (including distractor images), richer annotations (image-level and head-level in addition to point-level annotations). %The dataset is randomly split into 3,888 for training and 1,062 for testing, respectively.

\noindent $\bullet$ \textbf{DLR-ACD~\cite{bahmanyar2019mrcnet}}\footnote{https://www.dlr.de/eoc/en/desktopdefault.aspx/tabid-12760/22294\_read-58354/} contains 33 large aerial images with average resolution is 3619$\times$5226, which are captured by standard DSLR cameras installed on an airborne platform on a helicopter. The images come from 16 flight campaigns, and the dataset contains 226,291 person annotations in total. %The dataset is split 19 images as training and other 14 images as testing.

\noindent $\bullet$ \textbf{DroneCrowd~\cite{wen2019drone}} is a drone-based dataset for density map estimation, crowd localization and tracking, simultaneously. The dataset is composed of 112 video sequences with 33,600 frames in total. The average resolution of the frames is 1920$\times$1080, collected from multiple drone devices, 70 different scenarios across four different cities in China. There are more than 4.8 million head annotations on 20,800 people trajectories.

\noindent $\bullet$ \textbf{GCC~\cite{wang2019learning}}\footnote{https://gjy3035.github.io/GCC-CL/} is collected from an electronic game Grand Theft Auto V (GTA5), named as "GTA5 Crowd Counting" (GCC for short), which consists of 15,212 images, with resolution of 1080$\times$1920, containing 7,625,843 persons. Compared with the existing datasets, GCC has four advantages: 1) free collection and annotation; 2) larger data volume and higher resolution; 3) more diversified scenes, and 4) more accurate annotations.

\noindent $\bullet$ \textbf{NWPU-Crowd~\cite{wang2020nwpu}}\footnote{http://www.crowdbenchmark.com/} contains 5,109 images with 2,133,238 annotated instances in total. Compared with previous dataset in real world, in addition to data volume, there also some other advantages including negative samples, fair evaluation, higher resolution and large appearance variation.

\begin{table*}[!htb]
    \scriptsize
	\caption{Statistics of the object counting datasets, including crowd counting and other fields. Total-, min-, average- and max represent the total number, the minimum, average number and maximum number of instances in the datasets, respectively.}
	\vspace{-0.3cm}
	\begin{center}
		\renewcommand{\arraystretch}{1.0}	
		\setlength\tabcolsep{2pt}
    \begin{tabular}{r|c|c|c|c|c|c|c|c|c}

    \hline

    \multicolumn{1}{r|}{\multirow {2}{*}{Dataset}} &\multicolumn{1}{c|}{\multirow {2}{*}{Year}} &\multicolumn{1}{c|}{\multirow {2}{*}{Attributes}} &\multicolumn{1}{c|}{\multirow {2}{*}{Number of Images}} &\multicolumn{1}{c|}{\multirow {2}{*}{Training/Test}} &\multicolumn{1}{c|}{\multirow{2}{*}{Average Resolution}} &\multicolumn{2}{c}{Count Statistics} \\
    %\hline
    \cline{7-10}
    ~ &~ &~ &~ &~ &~  &Total &Min  &Average &Max \\
    \hline
    LHI$^1$~\cite{yao2007introduction} &2007 &Real-world &--  &-- &352 $\times$ 288 &-- &-- &--&-- \\
    UCSD~\cite{chan2008privacy} &2008 &Real-world & 2000  &800/1200 & 238 $\times$ 158 & 49,885 & 11 & 24.9 & 46 \\
    PETS~\cite{ellis2010pets2010} &2010 &Real-world &1076 &-- &384$\times$288 &18289 & 0 &-- & 40 \\
    Mall~\cite{chen2012feature} &2012 &Real-world & 2000 &800/1200 & 320 $\times$ 240 & 62,325 & 13 & 31 & 53 \\
    UCF\_CC\_50~\cite{idrees2013multi} &2013 &Real-world & 50 &-- & 2101 $\times$ 2888 & 63,974 & 94 & 1,280 & 4,543 \\
    AHU-Crowd~\cite{lim2014crowd} &2014 &Real-world &-- &-- &-- &-- & -- & -- & -- \\
    MICC~\cite{bondi2014real} &2014 &Real-world &3358 &-- &-- & 17630 & 0 & 5.25 & 28 \\
    WorldExpo'10~\cite{zhang2015cross} &2015 &Real-world  &3980 &-- &576 $\times$ 720 & 199,923 & 1 & 50.2 & 253 \\

    Indoor$^1$~\cite{luo2016real} &2016 &Real-world &570,000 &-- &704$\times$576 &-- &0 &-- &59 \\
    LHI$^2$~\cite{zhao2016crossing} &2016 &Real-world & 3,100 &-- &1280 $\times$ 720&5,900 & -- & -- & -- \\
    AHU-CROWD~\cite{hu2016dense}&2016 &Real-world & 107 &-- &--& 45,000 & 58 & -- & 2201 \\
    SHT\_A~\cite{zhang2016single} &2016 &Real-world  & 482 &300/182 & 589 $\times$ 868 & 241,677 & 33 & 501.4 & 3,139 \\
    SHT\_B~\cite{zhang2016single} &2016 &Real-world & 716 &400/316 & 768 $\times$ 1024 & 88,488 & 9 & 123.6 & 578 \\
    Train Station~\cite{farhood2017counting} &2017 &Real-world  & 2000 &-- & 256 $\times$ 256 & 62581 & 1 & -- & 53 \\
    STF(C5\&C9)~\cite{farhood2017counting}  &2017 &Real-world  & 788$\&$600 &-- & 576 $\times$ 704 & -- & 3 & -- & 65 \\
    Shanghai Subway Station~\cite{he2017double} &2017 &Real-world &3,000 &-- &-- &-- &28.78 &-- &-- \\
    Beijing BRT~\cite{ding2018deeply}  &2018 &Real-world  & 1280 &-- & 640 $\times$ 360 & -- & 1 & -- & 64 \\
    EBP~\cite{zheng2018cross} &2018 &Real-world  &-- &-- &720 $\times$ 408 &-- &-- &-- &-- \\
    Smartcity~\cite{zhang2018crowd}  &2018 &Real-world  &50 &-- &1920 $\times$ 1080 &369 &1 &7.4 &14 \\
    CrowdFlow~\cite{schroder2018optical} &2018 &Synthetic &-- &-- &300$\sim$450 &-- &-- &-- &--  \\
    UCF-QNRF~\cite{idrees2018composition} &2018 &Real-world  & 1,535 &1201/334 & 2013 $\times$ 2902 & 1,251,642 & 49 & 815 & 12,865 \\
    CIISR~\cite{xu2019depth} &2019 &Real-world  &1000 &-- &1080 $\times$ 720 &-- &-- &117 &-- \\
    Venice~\cite{liu2019context} &2019 &Real-world  &167 &-- &1280 $\times$ 720 &-- &-- &-- &-- \\
    Indoor$^2$~\cite{ling2019indoor} &2019 &Real-world  &148,243 &-- &352 $\times$ 288 or 704$\times$576 &1,834,770 &0 &12.4 &40 \\

    City Street~\cite{zhang2019wide} &2019 & Real-world  &500 &300/200 &676 $\times$ 380 &-- &70 &-- &150 \\
    ShanghaiTechRGBD~\cite{lian2019density} &2019 & Real-world &2193 &1193/1000 &1080 $\times$ 1920 &144,512 &6 &65.9 &234 \\
    FDST~\cite{fang2019locality} &2019 &Real-world &15,000 &9000/6000 &1920 $\times$ 1080 and 1280 $\times$ 720 &394,081 &9 &26.7 &57 \\
    Crowd Surveillance~\cite{yan2019perspective} &2019 &Real-world &13,945 &-- &1342 $\times$ 840 &386,513 &-- &35 &-- \\
    JHU-CROWD~\cite{sindagi2019pushing} &2019 &Real-world &4250 &3888/1062 &1450 $\times$ 900 &1,114,785 &-- &262 &7286 \\
    ZhengzhouAirport~\cite{jiang2019learning} &2019 &Real-world &1,111 &-- &-- &49,061 &7 &-- &128 \\
    DLR-ACD~\cite{bahmanyar2019mrcnet} &2019 &Aerial imagery &33 &19/14 &3619 $\times$ 5226 &226,291 &285 &6857 &24,368 \\
    DroneCrowd~\cite{wen2019drone} &2019 & Drone-based &33,600 &-- &1920 $\times$ 1080 &4,864,280 &25 &144.8 &455 \\
    Categorized~\cite{das2019cccnet} &2019 &Categories counting &553 &-- &-- &16,521 &1 &29.8 &206 \\
    GCC~\cite{wang2019learning} &2019 &Synthetic  & 15,212 &-- & 1080 $\times$ 1920 & 7,625,843 & 0 & 501 & 3,995 \\
    NWPU-Crowd~\cite{wang2020nwpu} &2020 &Real-world &5,109 &-- &2311 $\times$ 3383 &2,133,238 &0 &418 &20,033 \\

    \hline

    Caltech~\cite{dollar2012pedestrian} &2012 &Pedestrian detection  &2000 &-- &-- &15043 &6 &-- &14 \\
    TRANCOS~\cite{guerrero2015extremely} &2015 &Vehicle counting  &1244 &403/420/421 &640 $\times$ 480 &46,796 &9 &-- &107 \\
    Penguins~\cite{arteta2016counting}  &2016 & Penguins counting  &80095 &-- &-- &-- &0 &7.18 &67 \\
    WIDER FACE~\cite{yang2016wider} &2016 &Face detection &32,203 &40$\%$/10$\%$/50$\%$ &-- &393,703 &-- &-- &-- \\
    DukerMTMC~\cite{ristani2016performance} &2016 &tracking, human detection or ReID &over 2 million &-- &1920$\times$1080 &2,700 &-- &-- &-- \\
    WebCamT~\cite{zhang2017understanding} &2017 &WebCam traffic counting  &60 million &42,200/17800 &352$\times$240&-- &-- &-- &-- \\
    CARPK~\cite{hsieh2017drone} &2017 &Drone view-based car counting &1448 &989/459 &-- &89,777 &-- &-- &-- \\
    MTC~\cite{lu2017tasselnet} &2017 &Planting counting &361 &186/175 &-- &-- &-- &-- \\
    DCC~\cite{marsden2018people} &2018 &Cell counting  &177 &100/77 &-- &-- &0 &34.1 &101 \\
    Wheat-Spike~\cite{josuttes2018utilizing} &2018 &wheat spikes counting &20 &8/2/10 &1k$\sim$3k  &20,101 &749 &1005 &1287 \\

    VisDrone2019 People~\cite{zhu2018vision} &2018 &Drone-based crowd counting &3347 &2392/329/626 &969$\times$1482 &108,464 &10 &32.41 &289\\
    VisDrone2019 Veheicle~\cite{zhu2018vision} &2018 &Drone-based vehicle counting &5303 &3953/364/986 &991$\times$1511 &198,984 &10 &37.52 &349\\
    \hline
    \end{tabular}
    \label{tabel:datasets}
    \end{center}
\end{table*}

\subsection{Some special crowd counting datasets}
\label{subsection:some}
In this subsection, we briefly introduce some special crowd counting datasets, which are only used in some certain scenarios.
These datasets contain line crowd counting (LHI~\cite{yao2007introduction,zhao2016crossing}, crowd sequences (PETS~\cite{ellis2010pets2010}, Venice~\cite{liu2019context}), multi-sources (AHU-Crowd~\cite{lim2014crowd,hu2016dense}, CIISR~\cite{xu2019depth}, Venice~\cite{liu2019context}), indoor (MICC~\cite{bondi2014real}, Indoor$^1$~\cite{luo2016real}, Indoor$^2$~\cite{ling2019indoor}), train station (TS~\cite{farhood2017counting}, STF~\cite{farhood2017counting}), subway station (Shanghai Subway Station~\cite{he2017double}), BRT (Beijing BRT~\cite{ding2018deeply}\footnote{https://github.com/XMU-smartdsp/Beijing-BRT-dataset}), bridge (EBP~\cite{zheng2018cross}), airport (ZhengzhouAirport~\cite{jiang2019learning}), categorized~\cite{das2019cccnet}. The specific statistics of these datasets are listed in Tabel~\ref{tabel:datasets}.

\subsection{Representing object Counting datasets in other fields}
\label{subsection:representing}
For completeness, we introduce some representing object counting dataset in other fields, to verify the generalization ability of the designed model further.

\noindent $\bullet$ \textbf{Caltech~\cite{dollar2012pedestrian}} is a dataset for pedestrian detection, which has 2000 images and 15043 pedestrians in total. %The crowd varies between 6 and 14.
The dataset can be used to verify the performance of the algorithms in sparse scenes.

\noindent $\bullet$ \textbf{TRANCOS~\cite{guerrero2015extremely}}\footnote{http://agamenon.tsc.uah.es/Personales/rlopez/data/trancos} is the first one for vehicle counting in traffic jam images. %which consists of 1244 images with 46,796 vehicles annotated, where 403 images for training, 420 images for validation and 421 images for testing. All the images have the size of 640$\times$480.
The dataset is often used to evaluate the generalization ability of the crowd counting methods.

\noindent $\bullet$ \textbf{The penguin dataset}~\cite{arteta2016counting}\footnote{www.robots.ox.ac.uk/$~$vgg/research/penguins} is a product of an ongoing project for monitoring the penguin population in Antarctica.%which contains 500 thousand images with 80095 annotated, the count range is between 0 and 57. The image resolutions between 1MP and 6MP, and the sizes of penguins are ranging from 15 pixels to 700 pixels in length.
The dataset can be used to study climate change, etc.

\noindent $\bullet$ \textbf{WIDER FACE~\cite{yang2016wider}}\footnote{http://mmlab.ie.cuhk.edu.hk/projects/WIDERFace/} is a large-scale face detection benchmark, which is composed of 32,203 images in total and 393,703 faces with bounding boxes annotations. %It is constructed based on 60 event classes and divided in 40\% training, 10\% validation and 50\% testing.
%Compared with the datasets for crowd counting, WIDER FACE is more challenging due to high degree of variability in scale, pose, occlusion, expression, appearance and illumination.

\noindent $\bullet$ \textbf{DukerMTMC~\cite{ristani2016performance}} is a multi-view video dataset for multi-view tracking, human detection and re-identification (ReID), which contains over 2 million frames and more than 2,700 identities.% The video are taken from 8 synchronized cameras for 85 minutes with 1080p resolution at 60fps.

\noindent $\bullet$ \textbf{WebCamT~\cite{zhang2017understanding}} is the first largest annotated webcam traffic dataset to date, which consists of 60 million frames collected from 212 web cameras with different locations, camera perspective, and traffic states. %of which 60,000 images are annotated, 42,200 and 17,800 are used for training and testing set, respectively.
%In contrast to existing traffic datasets, the data are more challenging due to the low frame rate, low resolution, high resolution and large perspective.

\noindent $\bullet$ \textbf{CARPK~\cite{hsieh2017drone}}\footnote{https://lafi.github.io/LPN/} is a car counting datasets collected from 4 different parking lots with drone view, which contains nearly 90,000 cars in total, all the images with bounding box annotations. %989 training images are taken from three parking lot and 459 testing images are from the fourth parking lot.
%It is the first and the largest dataset that can support car counting in contrast to other parking lot datasets.

\noindent $\bullet$ \textbf{MTC~\cite{lu2017tasselnet}} includes 361 high-resolution images of maize tassels in the wild filed. Compared with the other objects that have similar physical sizes, maize tassels have the heterogeneous physical sizes and self-changing over time. Thus it is more suitable for evaluating the robustness to object-size variations of the designed model. %The dataset is split 186 images for training and validation and 175 images for test.

\noindent $\bullet$ \textbf{DCC~\cite{marsden2018people}} is a cell microscopy dataset, which contains 177 images with a variety of tissues and species. Across these images, the average count is 34.1, and the standard deviation is 21.8. %In addition, 100 images are used for training and validation whilst the rest 77 are served as test set.

\noindent $\bullet$ \textbf{Wheat-Spike~\cite{josuttes2018utilizing}} is a challenging dataset due to irregular placement or collection of wheat spikes, which contains 20 images in total, where the data are split into 8, 2 and 10 for training, validation and test.

\noindent $\bullet$ \textbf{VisDrone2019~\cite{zhu2018vision}} originates from an object detection dataset with bounding boxes annotated, Bai et al.~\cite{bai2019crowd} takes the center of bounding box as the location of objects, selecting pedestrian and people to form \textbf{VisDrone2019 People} dataset, and combining car, van, truck and bus to construct \textbf{VisDrone2019 Vehicle} dataset.

\section{Evaluation metrics}
\label{section:metrics}
There are several ways to evaluate the performance between predicted estimations and ground truths. In this section, we review some universally-agreed and popularly adopted measures for crowd counting model evaluation. According to different evaluation criteria, we divide the evaluation metrics into three categories: image-level for evaluating the counting performance, pixel-level for measuring the density map quality and point-level for assessing the precision of localization.

\subsection{Image-level metrics}
Two most common used metrics are Mean Absolute Error (MAE) and Mean Square Error (RMSE), which are defined as follows:
\begin{equation}
\footnotesize
MAE=\frac{1}{N} \sum_{i=1}^{N}\left|C_{I_{i}}^{p r e d}-C_{I_{i}}^{g t}\right|,
\end{equation}

\begin{equation}
\footnotesize
RMSE=\sqrt{\frac{1}{N} \sum_{i=1}^{N}\left|C_{I_{i}}^{p r e d}-C_{I_{i}}^{g t}\right|^{2}},
\end{equation}
where $N$ is the number of the test image, $C_{I_{i}}^{\text { pred }}$ and $C_{i}^{g t}$ represent the prediction results and ground truth, respectively. Roughly speaking, $MAE$ determines the accuracy of the estimates, while $RMSE$ indicates the robustness of the estimates.

Considering aforementioned $MAE$ may loss the location information, to provide a more accurate evaluation, Guerrero et al.~\cite{guerrero2015extremely} propose a new metric is Grid Average Mean Absolute Error (GAME), which is defined as follows:

\begin{equation}
\footnotesize
GAME(L)=\frac{1}{N} \sum_{n=1}^{N}\left(\sum_{l=1}^{4^{L}}\left|C_{I_{i}}^{p r e d}-C_{I_{i}}^{g t}\right|\right),
\end{equation}
where $4^{L}$ denotes that dividing the image into some non-overlapping regions. The higher $L$, the more restrictive of the GAME metric will be. Note that when $L=0$, $GAME$ will degenerate into $MAE$.

Similarly, accounting for the localization errors, a mean pixel-level absolute error (MPAE)~\cite{liu2019geometric} is proposed as follows:
\begin{equation}
\footnotesize
\mathrm{MPAE}=\frac{\sum_{i=1}^{N} \sum_{j=1}^{H} \sum_{k=1}^{W}\left|D_{i, j, k}-\hat{D}_{i, j, k}\right| \times 1_{\left\{D_{i, j, k} \in R_{i}\right\}}}{N},
\end{equation}
where $D_{i, j, k}$ denotes the ground-truth density map of $i$-th image at the pixel ($j$,$k$), $\hat{D}_{i, j, k}$ means the corresponding estimated density map, $R_{i}$ represents the ROI of the $i$-th image, $1_{\{\cdot\}}$ indicates the indicator function, and $W$, $H$ and $N$ are the width, height and the number of test samples. MPAE measures the degree of wrongly localized the densities are.

In view of both $MAE$ and $RMSE$ are the metrics for global accuracy and robustness, which cannot evaluate the local regions, thus Tian et al.~\cite{tian2019padnet} expand $MAE$ and $RMSE$ to patch mean absolute error ($PMAE$) and patch mean squared error ($PMSE$), which are defined as
\begin{equation}
\footnotesize
PMAE=\frac{1}{m \times N} \sum_{i=1}^{m \times N}\left|C_{I_{i}}^{p r e d}-C_{I_{i}}^{g t}\right|,
\end{equation}

\begin{equation}
\footnotesize
PMSE=\sqrt{\frac{1}{m \times N} \sum_{i=1}^{m \times N}\left(C_{I_{i}}^{p r e d}-C_{I_{i}}^{g t}\right)^{2}},
\end{equation}
where $m$ is the splitted non-overlapping patches. Note that when $m$ equals to 1, $PMAE$ and $PMSE$ degenerate into $MAE$ and $RMSE$, respectively.

\subsection{Pixel-level metrics}
Two metrics named Peak Signal to Noise Ratio (PSNR) and Structural Similarity Index (SSIM)~\cite{wang2004image} are usually used to measure the quality of the generated density map. Specifically, PSNR, the most common and widely used evaluation index of the image, which is essentially based on the error between corresponding pixels, in other words, error sensitivity. Generally speaking, high values represent smaller errors. However, it does not take the human visual characteristics into account, for example, the human is more sensitive to the contrast difference of lower spatial frequency and more sensitive to the brightness than hue, the perceptual results of a region is influenced by surrounding adjacent regions, etc. Therefore, the evaluation results are often inconsistent with people's subjective feelings.

%\begin{equation}
%PSNR=10 \log _{10}\left(\frac{\left(2^{n}-1\right)^{2}}{MSE}\right)
%\end{equation}
%where $n$ means the number of the bits per pixel, whose value equals 8 in general, corresponding to the gray-level which is 256. Generally speaking, larger values stands for smaller errors.

In addition, SSIM~\cite{wang2004image} measures the image similarity from three aspects: brightness, contrast and structure, which can be regarded as the multiplication of the three parts. The value of SSIM is in the range of [0,1], the larger of the value, the less distortion of the image.

%\begin{equation}
%l(X, Y)=\frac{2 \mu_{X} \mu_{Y}+C_{1}}{\mu_{X}^{2}+\mu_{Y}^{2}+C_{1}}
%\end{equation}
%
%\begin{equation}
%c(X, Y)=\frac{2 \sigma_{X} \sigma_{Y}+C_{2}}{\sigma_{X}^{2}+\sigma_{Y}^{2}+C_{2}}
%\end{equation}
%
%\begin{equation}
%s(X, Y)=\frac{\sigma_{X Y}+C_{3}}{\sigma_{X} \sigma_{Y}+C_{3}}
%\end{equation}
%where $C_1$, $C_2$ and $C_3$ are constants, to avoid zeros in the denominator, empirically, setting $C_1  = (K_1 *L)^2$, $C_2  = (K_2 *L)^2$, $C_3  = C_2/2$, and $K_1=0.01$, $K_2=0.03$ and $L=255$. Additionally, $\mathrm{U}_{\mathrm{X}}$, $\mathrm{U}_{\mathrm{Y}}$, $\sigma_{X}$ and $\sigma_{Y}$ represent the mean and variance of the images X and Y, and $\sigma_{X Y}$ indicates the covariance between image X and image Y, i.e.,
%
%\begin{equation}
%\mu_{X}=\frac{1}{H \times W} \sum_{i=1}^{H} \sum_{j=1}^{W} X(i, j)
%\end{equation}
%
%\begin{equation}
%\sigma_{X}^{2}=\frac{1}{H \times W-1} \sum_{i=1}^{H} \sum_{j=1}^{W}\left(X(i, j)-\mu_{X}\right)^{2}
%\end{equation}
%
%\begin{equation}
%\sigma_{X Y}=\frac{1}{H \times W-1} \sum_{i=1}^{H} \sum_{j=1}^{W}\left(\left(X(i, j)-\mu_{x}\right)\left(Y(i, j)-\mu_{y}\right)\right)
%\end{equation}
%
%And then,
%
%\begin{equation}
%SSIM(X, Y)=l(X, Y) \cdot c(X, Y) \cdot s(X, Y)
%\end{equation}
%The value of SSIM range in [0,1], larger values stand for smaller errors.

\subsection{point-level metrics}
To evaluate the localization performance of the model, Average Precision (AP) and Average Recall (AR) are two most common used metrics. Generally speaking, when the value of AP increases, AR decreases. Thus how to trade off between them is a worthy considering question.

\section{Benchmarking and analysis}
\label{section:benchmarking}

\subsection{Overall benchmarking results evaluation}
Table~\ref{table:benmarking} presents results of 53 state-of-the-art CNN-based methods and 7 representative traditional approaches over six mainstream benchmark datasets in crowd counting task. Two widely used evaluation metrics, i.e., MAE and RMSE are for measuring the accuracy and robustness of the models. All the models are representative and the results listed in the Table~\ref{table:benmarking} are published in their papers or reported by other works~\footnote{More detailed results leaderboard can be found at https://github.com/gjy3035/Awesome-Crowd-Counting}.

\noindent $\bullet$ \textbf{CNN-based v.s. traditional models.} Comparing the traditional models with CNN-based ones in Table~\ref{table:benmarking}, as excepted, we observe that CNN-based methods make great improvement of performances by a large margin. It also demonstrates that strong feature learning ability of deep convolution neural network based on large-scale annotated data.

\noindent $\bullet$ \textbf{Performance comparison of CNN-based models.} Since the year of 2015, the first CNN-based density map estimation model was proposed for crowd counting, the performance has also been improved gradually over time, which has witnessed the significant progress of the crowd counting model. Among the deep models, Cross scene~\cite{zhang2015cross} performs the worst performance, as the first ones to apply CNNs for crowd counting, which adopts the basic network structure and handle the cross-scene problem that transfers the pre-trained CNN to unseen scenes. Thus the model is lower compared with the single-scene and domain-specific model. However, this work provides a good solution to the generalizing trained model to unseen scenes.

\subsection{Properties-based evaluation}
We choose three top-performing models in terms of MAE and RMSE over six commonly used datasets, ending up with collecting 19 models, including two heuristic models, i.e., MCNN~\cite{zhang2016single} and CSRNet~\cite{li2018csrnet}, the main properties of these state-of-the-arts are listed in Table~\ref{table:properties}. These properties cover the main techniques that could be used to explain the reason they perform well.

From Table~\ref{table:properties}, we can find that, among these state-of-the-art methods, two-thirds of which adopt single column network architecture. For this phenomenon, perhaps we can reach the following conclusion: instead of making the network wider, deeper networks may be better. In addition, more than one third of them incorporate visual attention mechanism~\cite{liu2018crowd,liu2018decidenet,liu2019adcrowdnet,hossain2019crowd,varior2019scale,zhu2019dual,shi2019counting,sindagi2019ha-ccn} and dilation convolution layer~\cite{li2018csrnet,wu2019adaptive,chen2019scale,liu2019adcrowdnet,wang2019learning,liu2019context,dai2019dense} into their frameworks. Instead of using all available information of the input image in many CNNs-based methods, the visual attention mechanism is to use pertinent information to compute the neural responses, which can learn to weight the importance of each pixel of feature maps. Due to the prominent ability, visual attention mechanism has been applied to many computer vision tasks, such as image classification~\cite{hu2018squeeze}, semantic segmentation~\cite{ren2017end}, image deblurring~\cite{qian2018attentive}, and visual pose estimation~\cite{chu2017multi}, it is also suitable for the problem of crowd counting, highlighting the regions of interest containing the crowd and filtering out the noise in the background clutter situations. Dilated convolution layers, a good alternate of the pooling layer, have demonstrated that significant improvement of accuracy in segmentation tasks~\cite{yu2015multi,chen2018deeplab,chen2017rethinking}. The advantage of the dilation convolution layer is that enlarging receptive field without information loss caused by pooling operations (max and average pooling, etc.) and without increasing the number of parameters and the number of computations (such as up-sampling operations of the de-convolution layer in FCN~\cite{long2015fully}). Therefore, the dilation convolution layer can be integrated into the crowd counting framework to capture more multi-scale features and maintain more detailed information.

Spatial Transformer Network (STN)~\cite{jaderberg2015spatial} and deformable convolution~\cite{dai2017deformable} have a similar effect to address the problem of rotation, scaling or warping, which limit the capacity of feature invariance of standard CNNs. Specifically, STN is a sub-differential sampling module, which requires no extra annotations and has the capacity of adaptively learning spatial transformation between different data. It can not only carry out the spatial transformation on the input image but also any layer of the convolutional layer to realize the spatial transformation of different feature maps. Due to the remarkable performance, STN has been applied to many communities, such as multi-label image recognition~\cite{wang2017multi} and saliency detection~\cite{kuen2016recurrent}. Therefore, it also adopted by Liu et al.~\cite{liu2018crowd} to address the scale and rotation variation in crowd counting.

%Meanwhile, deformable convolution is an operation which adds an offset, whose magnitude can be learnable, on each point in the receptive field of feature map. After the offset, the shape of receptive field is matching the actual shape of the object rather than a square. The advantages of the offset is that no matter how deformable the objet is, the region of convolution always cover the object. Thereby when it is applied to crowd counting in ~\cite{liu2019adcrowdnet}, it can well cope with the distortion caused by the perspective view of the camera and diverse crowd distribution in real world.

Conditional random fields (CRFs)~\cite{lafferty2001conditional} or Markov Random Fields (MRFs)~\cite{li1994markov} have been usually leveraged as a post-possessing operation to refine the features and outputs of the CNNs with a message passing mechanism~\cite{krahenbuhl2011efficient}. In the work of ~\cite{liu2019crowd}, the first to utilize CRFs to refine features with different scales for the crowd counting task and demonstrates the effectiveness on the benchmark datasets. Zhang et al.~\cite{zhang2019attentional} propose an attentional neural fields (ANF) framework which integrates CRFs and non-local operation~\cite{wang2018non} (similar as self-attention) for crowd counting.

Perspective distortion is a major challenge in the crowd counting, while perspective information may be provided in two ways: one is related to camera's 6 degree-of-freedom (DOF)~\cite{gao2003complete}, the other is to identify the scale variation in the distance from the camera in the counting task. It can provide additional information with respect to scale variation and perspective geometry, many traditional crowd counting methods~\cite{chan2008privacy,chan2012counting} utilize it to normalize the regression features or detection features of changeable scales. Some modern CNNs-based methods also use perspective information to infer the ground truth density~\cite{zhang2015cross,zhang2016single} or body part maps~\cite{huang2018body}. These methods utilize perspective information yet without using the perspective map. Instead, some works~\cite{liu2019context,shi2019revisiting} leverage it to encode global or local scales in the network.

%The concept of scenarios is very abstract, which is difficult to define directly, hence many algorithms ignore the point, they simply assume that scenario is known, it cannot satisfy the situation in the real world. Fortunately, Wu et al.~\cite{wu2019adaptive} construct a adaptive scenario discovery (ASD) framework for crowd counting by combining two pathways corresponding to dense or sparse crowd whose concept is relative, through some discrete weights which are generated by adaptively learning, so that discover the scenarios implicitly.

Spatial pyramid pooling (SPP)~\cite{he2015spatial} was originally raised for visual recognition, which has several advantages than traditional networks, firstly it adapts the input images with arbitrary sizes; additionally, as the pooling layers with different sizes are extracted from the feature map and then aggregating them into a vector with fixed length, so that improve the robustness and accuracy. Moreover, it can accelerate convergence speed. Therefore, it is used to capture and fuse multi-scale features in SCNet~\cite{wang2018defense}, PaDNet~\cite{tian2019padnet} and CAN~\cite{liu2019context} for crowd counting.

Pan-density crowd counting aims to deal with two phenomena in crowd scenarios: varying densities and distributions in different scenarios and inconsistent densities of local regions in the same scene. Most current methods are designed for a specific density or scenario so that it is difficult to take full advantage of pan-density information. Although many multi-column architectures are designed to cope with this problem, such as MCNN~\cite{zhang2016single}, Switch-CNN~\cite{sam2017switching} and CP-CNN~\cite{sindagi2017generating}, they always suffer from low efficiency, high computation complexity, and biased local estimation. However, PaDNet~\cite{tian2019padnet} is put forward to provide a reasonable solution to effectively identify specific crowd by the sub-networks in Density-Aware Network (DAN), and learn an enhancement rate for each feature map by a Feature Enhancement Layer (FEL). In the final,  these feature maps are fused to obtain better counting.

\begin{table*}[!htb]
\scriptsize
	\caption{Comparison of the performance of different methods on the representing crowd counting datasets. \mbox{\color{red}{\textbf{Red}}}, \mbox{\color{green}{\textbf{green}}} and \mbox{\color{blue}{\textbf{blue}}} indicate the first, the second and the third best performance, respectively. Note that the MAE in WorldExpo'10~\cite{zhang2015cross} is the average value of the five cross-scenes, SFCN$\dag^2$~\cite{wang2019learning} represents that model takes ResNet101 as backbone, pre-trained on GCC~\cite{wang2019learning}, "--" denotes that results are not available and "$\downarrow$" indicates the lower the better of the results.}
	\vspace{-0.3cm}
	\begin{center}
		\renewcommand{\arraystretch}{1.0}
		\setlength\tabcolsep{1.8pt}
		\begin{tabular}{|c|r|l|lr|lr|lr|lr|lr|lr|lr|}
			\hline
             \multicolumn{1}{|c|}{\multirow {2}{*}{\#}} &\multicolumn{1}{c|}{\multirow {2}{*}{Methods}} &\multicolumn{1}{c|}{\multirow {2}{*}{Year$\&$Venue}} &\multicolumn{2}{c|}{UCSD~\cite{chan2008privacy}} &\multicolumn{2}{c|}{Mall~\cite{chen2012feature}} &\multicolumn{2}{c|}{UCF\_CC\_50~\cite{idrees2013multi}} &\multicolumn{2}{c|}{WorldExpo'10~\cite{zhang2015cross}} &\multicolumn{2}{c|}{SHT\_A~\cite{zhang2016single}} &\multicolumn{2}{c|}{SHT\_B~\cite{zhang2016single}} &\multicolumn{2}{c|}{UCF-QNRF~\cite{idrees2018composition}} \\
			\cline{4-17}
            ~ &~ &~ &MAE$\downarrow$ &RMSE$\downarrow$ &MAE$\downarrow$ &RMSE$\downarrow$ &MAE$\downarrow$ &RMSE$\downarrow$ &MAE$\downarrow$ &RMSE$\downarrow$ &MAE$\downarrow$ &RMSE$\downarrow$ &MAE$\downarrow$ &RMSE$\downarrow$ &MAE$\downarrow$ &RMSE$\downarrow$\\ \hline

             1 &GP~\cite{chan2008privacy} &2008 CVPR &2.24 &7.97 &3.72 &20.1 &-- &-- &-- &-- &-- &-- &-- &-- &-- &-- \\  \hline
             2 &Lempitsky et.al~\cite{lempitsky2010learning} &2010 NIPS &1.70 &-- &-- &-- &493.4 &487.1 &-- &-- &-- &-- &-- &-- &-- &-- \\ \hline
             3 &LBP+RR~\cite{chen2012feature} &2012 BMVC &-- &-- &-- &-- &-- &-- &31.0&-- &303.2 &371.0 &59.1 &81.7 &-- &-- \\ \hline
             4 &Idrees 2013~\cite{idrees2013multi} &2012 CVPR &-- &-- &-- &-- &468.0 &590.3 &-- &-- &-- &-- &-- &-- &315.0 &508.0 \\ \hline
             5 &CA-RR~\cite{chen2013cumulative} &2013 CVPR &2.07 &6.86 &3.43 &17.7 &-- &-- &-- &-- &-- &-- &-- &-- &-- &-- \\ \hline
             6 &Faster RCNN~\cite{ren2015faster} & 2015 NIPS &-- &-- &5.91 &6.60 &-- &-- &-- &--&-- &-- & 44.51 &53.22 &-- &-- \\ \hline
             7 &Count-Forest~\cite{pham2015count} &2015 ICCV &1.61 &4.40 &2.50 &10.0 &-- &-- &-- &-- &-- &-- &-- &-- &-- &-- \\ \hline \hline
             8 &Cross-scene~\cite{zhang2015cross} & 2015 CVPR &1.60 &3.31 &-- &-- &467.0 &498.5 &12.9 &-- &181.8 &277.7 &32.0 &49.8 &-- &--\\ \hline
             9 &MCNN~\cite{zhang2016single} & 2016 CVPR &1.07 &1.35 &2.24 &8.5 &377.6 &509.1 &11.6 &-- &110.2 & 173.2 &26.4 &41.3 &-- &--  \\ \hline
             10 &MSCNN~\cite{zeng2017multi} & 2017 ICIP &-- &-- &-- &-- &363.7 &468.4 &11.7 &-- & 83.8 & 127.4 &17.7 &30.2 &-- &-- \\ \hline
             11 &CMTL~\cite{sindagi2017cnn} & 2017 AVSS &-- &-- &-- &-- &322.8 &397.9 &-- &-- &101.3 &152.4 &20.0 &31.1 &252 &514 \\ \hline
             12 &Switching CNN~\cite{sam2017switching} & 2017 CVPR &1.62 &2.10 &-- &-- &318.1 &439.2 &9.4 &-- &90.4 &135.0 &21.6 &33.4 &228 &445\\ \hline
             13 &CP-CNN~\cite{sindagi2017generating} & 2017 ICCV &-- &-- &-- &-- &295.8 &320.9 &8.86 &-- &73.6 &106.4 &20.1 &30.1 &-- &-- \\ \hline
             14 &SaCNN~\cite{zhang2018crowd} & 2018 WACV &-- &-- &-- &-- &314.9 &424.8 &8.5 &-- & 86.8 & 139.2 &16.2 &25.8 &-- &-- \\ \hline
             15 &ACSCP~\cite{shen2018crowd} & 2018 CVPR &1.04 &1.35 &-- &-- &291.0 &404.6 &7.5 &-- &75.7 &102.7 &17.2 &27.4 &-- &-- \\ \hline
             16 &CSRNet~\cite{li2018csrnet} &2018 CVPR  &1.16 &1.47 &-- &-- &266.1 &397.5 &8.6 &-- &68.2 &115.0 &10.6 &16.0 &120.3 &208.5 \\ \hline
             17 &IG-CNN~\cite{babu2018divide} &2018 CVPR &-- &-- &-- &-- &291.4 &349.4 &11.3 &-- &72.5 &118.2 &13.6 &21.1 &-- &-- \\ \hline
             18 &DecideNet~\cite{liu2018decidenet} & 2018 CVPR &-- &-- &\color{green}{\textbf{1.52}} &\color{green}{\textbf{1.90}} &-- &-- &9.23 &-- &-- &-- &21.53 &31.98 &-- &-- \\ \hline
             19 &DRSAN~\cite{liu2018crowd} & 2018 IJCAI &-- &-- &\color{blue}{\textbf{1.72}} &\color{blue}{\textbf{2.1}} &219.2 &\color{blue}{\textbf{250.2}} &7.76 &-- &69.3 &96.4 &11.1 &18.2 &-- &-- \\ \hline
             20 &ic-CNN (two stages)~\cite{ranjan2018iterative} & 2018 ECCV &-- &-- &-- &-- &260.9 &365.5 &10.3 &-- &68.5 &116.2 &10.7 &16.0 &-- &-- \\ \hline
             21 &SANet~\cite{cao2018scale} & 2018 ECCV &1.02 &1.29 &-- &-- &258.4 &334.9 &8.2 &-- &67.0 &104.5 &8.4 &13.6 &-- &--\\ \hline
             22 &SCNet~\cite{wang2018defense} &2018 arXiv &-- &-- &-- &-- &280.5 &332.8 &\color{red}{\textbf{6.4}} &-- &71.9 &107.9 &9.3 &14.4 &-- &-- \\ \hline
             23 &MA-Net~\cite{jiang2019mask} &2019 TCSVT &-- &-- &1.76 &2.2 &245.4 &349.3 &8.34 &-- &61.8 &100.0 &8.6 &13.3 &--  &-- \\ \hline
             24 &PaDNet~\cite{tian2019padnet} &2019 TIP &\color{green}{\textbf{0.85}} &\color{green}{\textbf{1.06}} &-- &-- &\color{green}{\textbf{185.8}} &278.3 &-- &-- &\color{blue}{\textbf{59.2}} &98.1 &8.1 &12.2 &\color{blue}{\textbf{96.5}} &170.2 \\ \hline
             25 &ASD~\cite{wu2019adaptive} & 2019 ICASSP &-- &-- &-- &-- &196.2 &270.9 &-- &-- &65.6 &98.0 &8.5 &13.7 &-- &-- \\ \hline
             26 &SAA-Net~\cite{varior2019scale} & 2019 CVPR &-- &-- &-- &-- &238.2 &310.8 &-- &-- &63.7 &104.1 &8.2 &12.7 &97.5 &167.8 \\ \hline
             27 &PACNN~\cite{shi2019revisiting} &2019 CVPR &\color{blue}{\textbf{0.89}} &1.18 &-- &-- &267.9 &357.8 &7.8 &-- &66.3 &106.4 &8.9 &13.5 &-- &-- \\ \hline
             28 &SFCN$\dag^2$~\cite{wang2019learning} &2019 CVPR &-- &-- &-- &-- &214.2 &318.2 &-- &-- &64.8 &107.5 &7.6 &13.0 &102.0 &171.4 \\ \hline
             29 &CAN(ECAN)~\cite{liu2019context} &2019 CVPR &-- &-- &-- &-- &212.2 &\color{green}{\textbf{243.7}} &7.4 (\color{blue}{\textbf{7.2}}) &-- &62.3 &100.0 &7.8 &12.2 &107 &183 \\ \hline
             30 &SFANet~\cite{zhu2019dual} &2019 arXiv &\color{red}{\textbf{0.82}} &\color{blue}{\textbf{1.07}} &-- &-- &219.6 &316.2 &-- &-- &59.8 &99.3 &6.9 &10.9 &100.8 &174.5 \\ \hline
             31 &CFF~\cite{shi2019counting} &2019 ICCV &-- &-- &-- &-- &-- &-- &-- &-- &65.2 &109.4 &7.2 &12.2  &-- &--\\ \hline
             32 &DUBNet~\cite{oh2019crowd} &2019 AAAI &1.03 &1.24 &-- &-- &235.2 &332.7 &-- &-- &66.4 &111.1 &9.4 &15.1 &116 &178 \\ \hline
             33 &CTN~\cite{viresh2019crowd} &2019 arXiv &-- &-- &-- &-- &219.3 &331.0 &-- &-- &64.3 &107.0 &8.6 &14.6 &-- &-- \\ \hline
             34 &DENet~\cite{liu2019denet} &2019 arXiv &1.05 &1.31 &-- &-- &241.9 &345.4 &8.2 &-- &65.5 &101.2 &9.6 &15.4 &-- &-- \\ \hline
             35 &W-Net~\cite{valloli2019w} &2019 arXiv &\color{red}{\textbf{0.82}} &\color{red}{\textbf{1.05}} &-- &-- &201.9 &309.2 &-- &--&59.5 &97.3 &6.9 &\color{green}{\textbf{10.3}} &-- &-- \\ \hline
             36 &DSNet~\cite{dai2019dense} &2019 arXiv &\color{red}{\textbf{0.82}} &\color{green}{\textbf{1.06}} &-- &-- &\color{red}{\textbf{183.3}} &\color{red}{\textbf{240.6}} &-- &-- &61.7 &102.6 &\color{green}{\textbf{6.7}} &10.5 &\color{green}{\textbf{91.4}} &\color{blue}{\textbf{160.4}} \\ \hline
             37 &SAAN~\cite{hossain2019crowd} & 2019 WACV &-- &-- &\color{red}{\textbf{1.28}} &\color{red}{\textbf{1.68}} &271.6 &391.0 &-- &-- &-- &-- &16.86 &28.41 &-- &-- \\ \hline
             38 &ADCrowdNet~\cite{liu2019adcrowdnet} &2019 CVPR &1.09 &1.35 &-- &-- &273.6 &362.0 &7.3 &--&70.9 &115.2 &7.7 &12.9 &-- &-- \\ \hline
             39 &PSDDN~\cite{liu2019point} &2019 CVPR &-- &-- &-- &-- &359.4 &514.8 &-- &-- &85.4 &159.2 &16.1 &27.9 &-- &-- \\ \hline
             40 &TEDnet~\cite{jiang2019crowd} &2019 CVPR &-- &-- &-- &-- &249.4 &354.5 &8.0 &-- &64.2 &109.1 &8.2 &12.8 &113 &188 \\ \hline
             41 &SPN~\cite{chen2019scale} &2019 WACV &1.03 &1.32 &-- &-- &259.2 &335.9 &-- &-- &61.7 &99.5 &9.4 &14.4 &-- &-- \\ \hline
             42 &PCC Net~\cite{gao2019perspective} &2019 TCSVT &-- &-- &-- &-- &240.0 &315.5 &9.5 &-- &73.5 &124.0 &11.0 &19.0 &132 &191 \\ \hline
             43 &RAZ-Net~\cite{liu2019recurrent} &2019 CVPR &-- &-- &-- &-- &-- &-- &8.0 &-- &65.1 &106.7 &8.4 &14.1 &116 &195 \\ \hline
             44 &RReg(MCNN)~\cite{wan2019residual} &2019 CVPR &-- &-- &-- &-- &-- &-- &8.7 &-- &72.6 &114.3 &15.5 &23.1 &-- &-- \\ \hline
             45 &RReg(CSRNet)~\cite{wan2019residual} &2019 CVPR &-- &-- &-- &-- &-- &-- &8.5 &-- &63.1 &96.2 &8.72 &13.56 &-- &-- \\ \hline
             46 &AT-CFCN~\cite{zhao2019leveraging} &2019 CVPR &-- &-- &2.28 &2.90 &-- &-- &-- &-- &-- &-- &11.05 &19.66 &-- &--  \\ \hline
             47 &AT-CSRNet~\cite{zhao2019leveraging} &2019 CVPR &-- &-- &-- &-- &-- &-- &7.8 &-- &-- &-- &8.11 &13.53 &-- &-- \\ \hline
             48 &IA-DCCN~\cite{sindagi2019inverse} &2019 CVPR &-- &-- &-- &-- &264.2 &394.4 &-- &-- &66.9 &108.4 &10.2 &16.0 &125.3 &185.7  \\ \hline
             49 &HA-CCN~\cite{sindagi2019ha-ccn} &2019 TIP &-- &-- &-- &-- &256.2 &348.4 &-- &-- &62.9 &94.9 &8.1 &13.4 &118.1 &180.4  \\ \hline
             50 &L2SM~\cite{xu2019learn} &2019 ICCV &-- &-- &-- &-- &\color{blue}{\textbf{188.4}} &315.3 &-- &-- &64.2 &98.4 &7.2 &11.1 &104.7 &173.6 \\ \hline
             51 &DSSINet~\cite{liu2019crowd} &2019 ICCV &-- &-- &-- &-- &216.9 &302.4 &\color{green}{\textbf{6.67}} &-- &60.63 &96.04 &\color{blue}{\textbf{6.85}} &\color{blue}{\textbf{10.34}} &99.1 &\color{green}{\textbf{159.2}} \\ \hline
             52 &BL~\cite{ma2019bayesian} &2019 ICCV &-- &-- &-- &-- &229.3 &308.2 &-- &-- &62.8 &101.8 &7.7 &12.7 &\color{red}{\textbf{88.7}} &\color{red}{\textbf{154.8}} \\ \hline
             53 &LSC-CNN~\cite{sam2019locate} &2019 ICCV &-- &-- &-- &-- &225.6 &302.7 &8.0 &-- &66.4 &117.0 &8.1 &12.7 &120.5 &218.2 \\ \hline
             54 &SANet~\cite{cao2018scale}+SPANet~\cite{cheng2019learning} &2019 ICCV &1.00 &1.28 &-- &-- &232.6 &311.7 &7.7 &-- &59.4 &\color{green}{\textbf{92.5}} &\color{red}{\textbf{6.5}} &\color{red}{\textbf{9.9}} &-- &--\\ \hline
             55 &MBTTBF-SCFB~\cite{sindagi2019multi} &2019 ICCV &-- &-- &-- &-- &233.1 &300.9 &-- &-- &60.2 &\color{blue}{\textbf{94.1}} &8.0 &15.5 &97.5 &165.2 \\ \hline
             56 &S-DCNet~\cite{xiong2019SDCNet} &2019 ICCV &-- &-- &-- &-- &204.2 &301.3 &-- &-- &\color{green}{\textbf{58.3}} &95.0 &\color{green}{\textbf{6.7}} &10.7 &104.4 &176.1 \\ \hline
             57 &PGCNet~\cite{yan2019perspective} &2019 ICCV &-- &-- &-- &-- &259.4 &317.6 &8.1 &-- &\color{red}{\textbf{57.0}} &\color{red}{\textbf{86.0}} &8.8 &13.7 &-- &-- \\ \hline
             58 &ANF~\cite{zhang2019attentional} &2019 ICCV &-- &-- &-- &-- &250.2 &340.0 &8.1 &-- &63.9 &99.4 &8.3 &13.2 &110 &174 \\ \hline
             59 &RANet~\cite{zhang2019relational} &2019 ICCV &-- &-- &-- &-- &239.8 &319.4 &-- &-- &59.4 &102.0 &7.9 &12.9 &111 &190 \\ \hline
             60 &ACSPNet~\cite{ma2019atrous} &2019 Neurocomputing &1.02 &1.28 &1.76 &2.24 &275.2 &383.7 &9.8 &-- &85.2 &137.1 &15.4 &23.1 &-- &-- \\ \hline
             %61 &ACM-CNN~\cite{zou2019attend} &2019 Neurocomputing &1.01 &1.29 &2.3 &3.1 &291.6 &337 &8.56 &-- &72.2 &103.5 &17.5 &22.7 &-- &-- \\ \hline
             %62 &DDCN~\cite{wang2019removing} &2019 Neurocomputing &-- &-- &-- &-- &286.2 &479.6 &9.8 &-- &71.5 &110.4 &13.8 &20.1 &-- &-- \\ \hline
		\end{tabular}
    \label{table:benmarking}
	\end{center}
	\vspace{-0.4cm}
\end{table*}

%%%

\begin{table*}[!htb]

\caption{Main properties of state-of-the-art methods.}
\begin{center}
\begin{tabular}{|r|p{1cm}|p{1cm}|p{1cm}|p{1.0cm}|p{1.0cm}|p{1.3cm}|p{1.3cm}|p{1.0cm}|p{1.0cm}|p{1.3cm}|}
\hline
\diagbox{Methods}{Properties} &Multi-column &Single-column &Attention-based &Dilation convolution &Spatial transformer &CRFs/MRF &Perspective information &Pyramid pooling &Pan-density /sub-region \\ \hline
MCNN~\cite{zhang2016single} &\checkmark & & & & & &\checkmark  & &\checkmark\\ \hline
CSRNet~\cite{li2018csrnet} & &\checkmark & &\checkmark & & &  & &\\ \hline
DRSAN~\cite{liu2018crowd} &\checkmark & &\checkmark & &\checkmark & &  & &\\\hline
DecideNet~\cite{liu2018decidenet} &\checkmark & &\checkmark & & & &  & & \\ \hline
SCNet~\cite{wang2018defense} & &\checkmark & &\checkmark & & &  &\checkmark &\\ \hline
PaDNet~\cite{tian2019padnet} &\checkmark & & & & & &  &\checkmark &\checkmark\\ \hline
%ASD~\cite{wu2019adaptive} &\checkmark & & &\checkmark & & & &\checkmark & &\\ \hline
%SPN~\cite{chen2019scale} & &\checkmark & &\checkmark & & & & & &\\ \hline
%ADCrowdnet~\cite{liu2019adcrowdnet} & &\checkmark &\checkmark &\checkmark & &\checkmark & & & &\\ \hline
SAAN~\cite{hossain2019crowd} &\checkmark & &\checkmark & & & &  & &\\ \hline
%SAA-Net~\cite{varior2019scale} & &\checkmark &\checkmark & & & & & & &\\ \hline
%SFCN$\dag^2$~\cite{wang2019learning} & &\checkmark & &\checkmark & & & & & &\\ \hline
PACNN~\cite{shi2019revisiting} & &\checkmark & & & & &\checkmark  & &\\ \hline
CAN\&ECAN~\cite{liu2019context} & &\checkmark & &\checkmark & & &\checkmark  &\checkmark &\\ \hline
SFANet~\cite{zhu2019dual} & &\checkmark &\checkmark & & & &  & &\\ \hline
%CFF~\cite{shi2019counting} & &\checkmark &\checkmark & & & & & & & \\ \hline
W-Net~\cite{valloli2019w} & &\checkmark &\checkmark & & & &  & &\\ \hline
DSNet~\cite{dai2019dense} & &\checkmark & &\checkmark  & & & & &\\ \hline
%HA-CCN~\cite{sindagi2019ha-ccn} & &\checkmark &\checkmark &  & & & & & &\\ \hline
L2SM~\cite{xu2019learn} & &\checkmark & &  & & & & &\\ \hline
DSSINet~\cite{liu2019crowd} &\checkmark & &\checkmark &\checkmark & &\checkmark & & &\\ \hline
SPANet~\cite{cheng2019learning} &\checkmark & & &  & & & & &\checkmark\\ \hline
MBTTBF-SCFB~\cite{sindagi2019multi} &\checkmark & &\checkmark & & &\checkmark & & &\\ \hline
BL~\cite{ma2019bayesian} & &\checkmark & &  & & & & &\\ \hline
S-DCNet~\cite{xiong2019SDCNet} & &\checkmark & &  & & & & &\checkmark\\ \hline
PGCNet~\cite{yan2019perspective} & &\checkmark & &  & & &\checkmark & &\\ \hline
\end{tabular}
\end{center}
\label{table:properties}
\end{table*}

Comprehensively considering the statistical results from Table~\ref{table:properties} and the analysis of various methods, we make the following observations:

\noindent $\bullet$ Among the CNN methods, most networks are based on single column network architecture, which is more simpler yet effective than multi-column architectures that are high complexity and bloated structure as demonstrated in ~\cite{li2018csrnet}.

\noindent $\bullet$ The techniques of visual attention mechanism, dilation convolution, and spatial pyramid pooling (SPP) can significantly improve the performance of the final estimation and the quality of density maps.

\noindent $\bullet$ Incorporating perspective information~\cite{zhang2015cross,zhang2016single,liu2019context,shi2019revisiting} into the network can provide additional support and guidance for the extraction of multi-scale features.

\noindent $\bullet$ Spatial transformer network~\cite{jaderberg2015spatial,liu2018crowd} and deformable convolution~\cite{dai2017deformable,liu2019adcrowdnet} can help to address the rotation and uniform distributions of crowds which is more suitable for the crowd understanding problem in the congested noisy scenarios.

\noindent $\bullet$ Pan-density learning~\cite{tian2019padnet} can not only take full advantages of global features but also make up biased local estimation.

\noindent $\bullet$ Multi-pathway or multi-task framework~\cite{liu2018decidenet,wu2019adaptive,zhu2019dual} followed by jointly loss function can improve the estimation performance and speed up the training.

\subsection{Attributes-based analysis}
\begin{figure*}[t]
%%tr = 0.006, ts = 0.008
  \centering
      \centerline{\includegraphics[width=1.0 \linewidth]{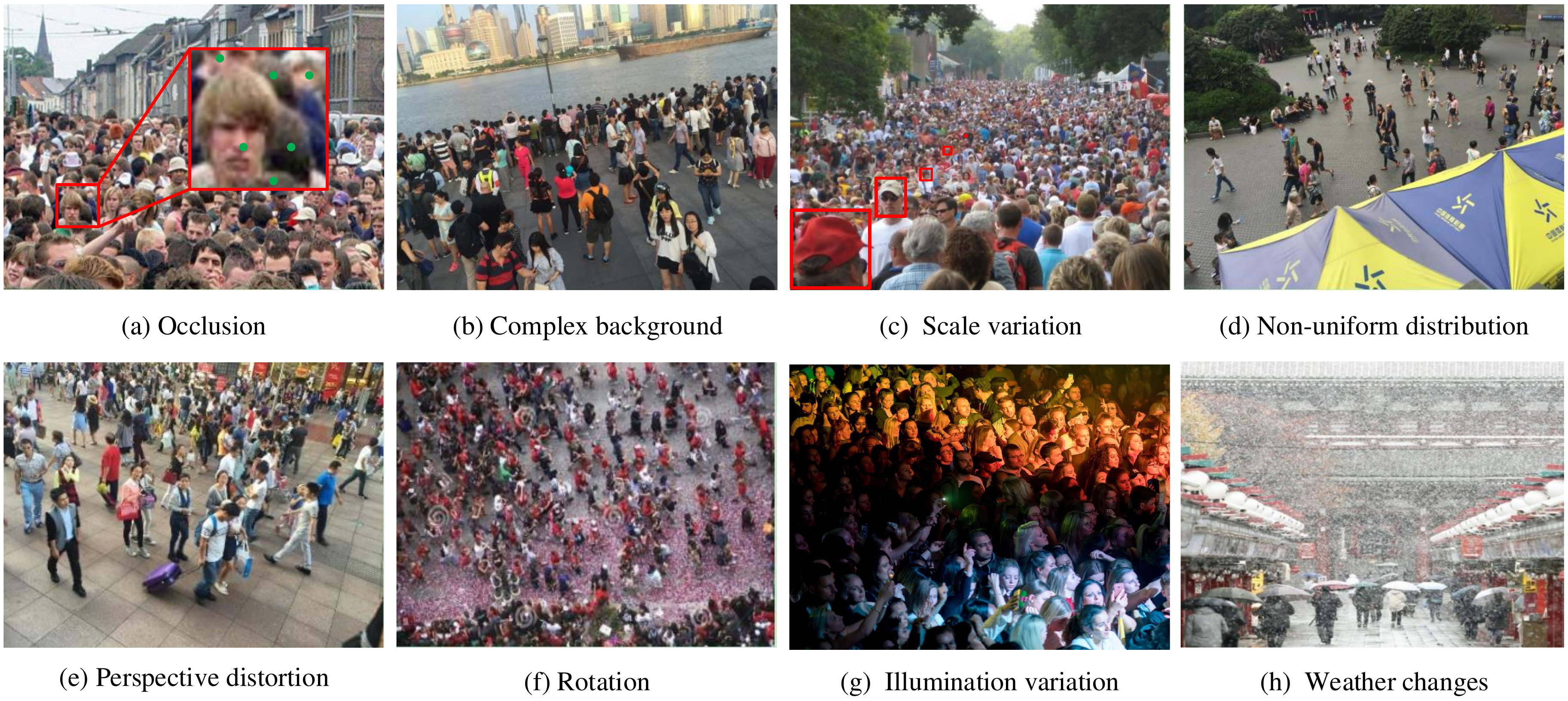}}
\caption{The challenges in crowd counting. }
\vspace{-5pt}
\label{fig:challenges}
\end{figure*}

Albeit significant performance improvement has been achieved to a great extent by applying CNNs into density map estimation crowd counting models, there are remain some challenges to be conquered. A robust network should have the capability of coping with various complex scenarios. The existence of challenges always brings many difficulties to the models, such as occlusion, scale variation, perspective distortion, rotation, illumination variation, and weather changes. Some samples are shown in Fig.~\ref{fig:challenges}. Moreover, the scenes of the images are from indoor, outdoor, and in the wild. It is worth noting that these attributes are not mutually exclusive. In other words, there may exist several attributes in one image. We also take the aforementioned methods in Table~\ref{table:properties} as an example to analyze these attributes in detail.

\noindent $\bullet$ \textbf{Occlusion}. As the crowd density increasing, the crowd will appear to occlude each other partly, which limits the capacity ability of traditional detection algorithms and prompts the emergence of density estimation models.

\noindent $\bullet$ \textbf{Complex background.} Background regions (have no person instances) includes confusing objects or have similar appearance or colors with the foreground, this can be suppressed through semantic segmentation or visual attention operations, such as~\cite{liu2018crowd,liu2018decidenet,liu2019adcrowdnet,hossain2019crowd,varior2019scale,zhu2019dual,chen2019crowd}.

\noindent $\bullet$ \textbf{Scale variation.} The most primary problem should be addressed in the density estimation models, as the scales of objects (such as the sizes of people heads) vary as the distance from the camera. Therefore, almost all the density estimation models are designed for addressing the scale variation problem in the first step.

\noindent $\bullet$ \textbf{Non-uniform distribution.} For intuitive understanding in the examples of Fig.~\ref{fig:challenges}, we can observe that diverse densities and distributions in different scenes and inconsistent distributions of local regions even in the same scene. The problem can be effectively addressed in the work of ~\cite{tan2019crowd}, which presents the first model to estimate various crowd densities with different regressors. Jiang et al.~\cite{jiang2019learning} tackle this issue by proposing a multi-level convolution neural network (MLCNN) which fuses multiple density maps generated by multi-level features. This problem can also be regarded as pan-density crowd counting, and related solutions can be referred to PaDNet~\cite{tian2019padnet}.

\noindent $\bullet$ \textbf{Perspective distortion.} Perspective distortion would drastically lead to person scale variation in the crowding counting scenes, which is related to camera calibration to estimate the 6 degree-of-freedom (DOF) of a camera.

\noindent $\bullet$ \textbf{Rotation.} The issue of rotation variation drastically due to the camera viewpoints such as different pose and photographic angles, it is addressed by the work ~\cite{liu2018crowd} via incorporating spatial transformer network (STN) into LSTM framework.

\noindent $\bullet$ \textbf{Illumination variation.} The illumination varies at different times in a day, usually from dark to light and then to dark, from dawn to dusk.

\noindent $\bullet$ \textbf{Weather changes.} The scenes in the wild usually under various types of weather conditions, such as clear, clouds, rain, foggy, thunder, overcast, and extra sunny.

The above challenges promote us to design more effective and robust frameworks to address these issues, and this also indicates that there is still much research room in the direction of crowd counting.

%%%%----------------------------------------------------------------------------------------------------------------------------------------------------------------------------------%%%%
\section{Discussion}
\label{section:discussion}

In this section, we discuss some important factors that will directly affect the performance of the crowd counting model design and some promising research directions.

\subsection{Model design}

\noindent $\bullet$ \textbf{Ground truth density maps generation.}
As a cornerstone towards CNN-based density estimation and crowd counting models, the generation of high-fidelity ground truth density maps is essential to data preparation for training. To convert an image with the original labels (generally refer to head location) to a density map, Lempitsky et al.~\cite{lempitsky2010learning} first raised and defined as a sum of Gaussian kernels centered on the locations of objects. This strategy works well for characterizing the density distribution of circle-like objects such as cells and bacteria. To tackle scale variations, Zhang et al.~\cite{zhang2015cross} put forward a solution by exploiting perspective information: the density map can be obtained by a sum of Gaussian kernels as a head part and a bivariate normal distribution. However, this strategy introduces a new issue of acquiring the perspective map.
Fortunately, Zhang et al.~\cite{zhang2016single} find that head size is related to the distance between two neighboring persons. Based on it, a geometry-adaptive kernel-based density map generation method is created, which inspires lots of works adopting this tool to prepare their data training. Although such strategy works well in the dense crowd scenes, it would fail in the sparse scenes. A depth-adaptive kernel-based density map generation method~\cite{lian2019density} is proposed by positing that the sizes of all heads are the same in the real world.

However, all the methods above are not content-aware. Therefore, Oghaz et al.~\cite{oghaz2019content} propose a brute-force nearest neighbor search technique to provide the absolute nearest neighbors despite the distribution of points, through using an integration of Chan-Vese segmentation algorithm, two-dimension Gaussian filter and brute-force nearest neighbor search technique. Recently, Xu et al.~\cite{xu2019autoscale} claim that the target density map generation manner may fail in the dense regions, since the density map is given by the sum of severely overlapped Gaussian blobs, which leads to diverse density patterns different from the sparse regions. Therefore, a learning-to-scale module (L2SM) is applied to re-scale the dense regions into similar scale levels, so as to ameliorate the pattern shifts and increase the counting accuracy. In another way, Ma et al.~\cite{ma2019bayesian} propose a Bayesian loss to enforce a more reliable density contribution probability model from the point-annotations.
Olmschenk et al.~\cite{olmschenk2019improving} propose an inverse k-nearest neighbor (ikNN) map to supersede the density maps, which can offer a smooth training gradient and accurate localization simultaneously, this also breaks the routine and opens new paths for us. Wan et al.~\cite{wan2019adaptive} propose an adaptive density map generator that generates a learnable density map representation from the ground truth dot labels.

Anyway, the proper selecting for density map generation will lay a solid foundation for the crowd counting.

\noindent $\bullet$ \textbf{Loss function.} The customized design of loss function is also an essential procedure in training effective models. Density map estimation CNN-based crowd counting methods are mostly a regression task, which usually adopt Euclidean distance as loss function to measure the difference between the estimated density map and ground truth. Although widely used, only the Euclidean loss employed may have some disadvantages such as sensitivity to outliers and image blur, pixel independent assumption neglecting the local coherence, and spatial correlation in density maps. Therefore, SmoothL1 loss~\cite{girshick2015fast} and Tukey Loss~\cite{belagiannis2015robust} are more robust for outliers can be leveraged, as described in ~\cite{zhang2019nonlinear}. Besides, an adversarial loss~\cite{goodfellow2014generative} is integrated to address the issue and improve the quality of density maps~\cite{sindagi2017generating,shen2018crowd,yang2018multi}. Nevertheless, density maps may contain little high-level semantic information, thus a light-weight SSIM local pattern consistency loss combined with Euclidean loss to enforce the local structural similarity between estimated density maps and ground truths~\cite{cao2018scale}, but it only can tackle the local consistent of regions with a fixed size. Therefore, further, a Dilated Multi-scale Structure Similarity (DMS-SSIM) loss~\cite{liu2019crowd,qiu2019crowd} is utilized to make the network learn the local similarity within the regions of varied sizes and generate the density maps with local consistency. Besides, a novel scale-aware loss function to specialize person head on a particular scale~\cite{varior2019scale}. Additionally, a combination of spatial abstraction loss (SAL) and the spatial correlation loss (SCL) are provided in~\cite{jiang2019crowd} to improve density map quality. In another way, accounting for the spatial variation of density, a Maximum Excess over Pixels (MEP) loss~\cite{cheng2019learning} is proposed to optimize the pixel-level subregion which has a high discrepancy with the ground-truth density map.

Overall, designing appropriate loss functions helps boost the performance of the models.

\noindent $\bullet$ \textbf{Information fusion of multi cues.} Generally, the information fusion of multiple cues can significantly improve the performance of the algorithm, for instance, the integration of scale-aware and context-aware would boost the performance~\cite{liu2019context,sindagi2017generating}, the combination of different pathways for sparse and dense scenarios~\cite{liu2018decidenet,wu2019adaptive,zhu2019dual}. Meanwhile, heterogeneous attributes, such as geometric/semantic/numeric attributes are leveraged to assist the density estimation for crowd counting~\cite{zhao2019leveraging}.

As a whole, there is an abundant of data availability across many different data sources or modalities in various formats. Fusing these heterogeneous cues with "broad learning" may be a reliable research direction.

\noindent $\bullet$ \textbf{Network Topology.} The network topology represents the information flow within the network, which influences the training complexity and parameters that are required. Proved by many experiments, the encoder-decoder pipeline appears promising performance for the crowd counting task, for instance, CSRNet~\cite{li2018csrnet} adopts a standard encoder-decoder structure, which takes a pre-trained VGG16~\cite{simonyan2014very} as a backbone, and builds dilated convolution operation in the decoder. SA-Net~\cite{cao2018scale} presents a similar model, which uses Inception model~\cite{szegedy2015going} in the encoder and Transposed convolution layers in the decoder. W-Net~\cite{valloli2019w} directly leverages U-Net~\cite{ronneberger2015u} structure with VGG16~\cite{simonyan2014very} replacing the encoder block, and adds a extra branch for faster convergence. TEDnet~\cite{jiang2019crowd} deploys an encoder-decoder hierarchy in a trellis manner. SGANet~\cite{wang2019segmentation} investigates the effectiveness of Inception-v3~\cite{szegedy2016rethinking} for crowd counting.
Beyond encoder-decoder pipeline, VGG16~\cite{simonyan2014very} is the best backbone for feature extraction among VGG16bn, Resnet50~\cite{he2016deep} and Inception~\cite{szegedy2015going}, which has been empirically demonstrated in~\cite{valloli2019w} and strong transfer ability for crowd analysis.

\subsection{Dataset construction}
In light of previous observation, we would put forward some suggestions for the construction of crowd counting dataset, w.r.t., scene diversity, multi-view, annotation accuracy, etc.

\noindent $\bullet$ \textbf{Scene diversity.} Some earlier datasets for crowd counting is in single-scene, i.e., the images are from the same video sequence, which has no variation in the perspective across different images, such as UCSD~\cite{chan2008privacy} and Mall~\cite{chen2012feature}, as illustrated in Fig.~\ref{fig:datasets}.

To meet the need of cross-scene and diversified data for deep model training, some more challenging datasets are proposed, including UCF\_CC\_50~\cite{idrees2013multi}, SHT\_A~\cite{zhang2016single}, UCF\_QNRF~\cite{idrees2018composition}, and countless others. Nevertheless, there are some drawbacks that limit their generation ability, for instance, UCF\_CC\_50~\cite{idrees2013multi} is limited by the small number of availability of high-resolution crowd images, SHT\_A~\cite{zhang2016single} is suffered from non-uniform density level and inaccurate labels of some samples. UCF\_QNRF~\cite{idrees2018composition} has the most number of high-count crowd images and annotations with diverse densities, which is more significant than UCF\_CC\_50~\cite{idrees2013multi}, however, the intra-class variation may sometimes exceed the capability of the network. Notably, they may tackle some unseen extreme cases when transferring the data to the wild, such as weather change and illumination variation. GCC~\cite{wang2019learning} may provide a reasonable solution by constructing a large-scale synthetic crowd counting, which consists of more diverse scenes to mimic the challenges in the wild better. Albeit plenty of data in GCC~\cite{wang2019learning}, there exists a large ``domain gap'' between synthetic and real data.

\noindent $\bullet$ \textbf{Multi-view.} Previous datasets are for single-view counting, which cannot satisfy the requirements of large and wide scenes, taking public parks or long queue in train station as an example, the scene is so wide that cannot be fully captured by single view, so long that too low of the resolution away from the camera, or most of the crowds are occluded by large objects. Some attempts have been made to tackle the shortcoming, for example, City street dataset~\cite{zhang2019wide} is collected from a busy street intersection, which contains large range crowds with more complex occlusion patterns and large scale variations.

\noindent $\bullet$ \textbf{Annotation accuracy.} There exists an intrinsic shortcoming in the existing dense crowd counting dataset, the annotations are not very accurate, such as some samples are from UCF\_CC\_50~\cite{idrees2013multi} and Shanghai Tech Part\_A~\cite{zhang2016single}. In fact, this problem is inevitable due to data annotated by different subjects or following different standards.

\noindent $\bullet$ \textbf{Annotation tools.} Effective annotation tools are critical in the process of dataset construction. We strongly recommend an online efficient annotation tool based on HTML5 + Javascript + Python, which supports two types of label form, i.e., point and bounding box. During the process of annotation, images are adaptively zoomed in/out to annotate heads according to different scales, and each of them are divided into $16\times16$ blocks, which provides five scales for the annotators, specifically $2^(i)$ ($i=0,1,2,3,4$) times size of the original image. This annotation tool can effectively improve the annotation speed and quality. With the aid of this effective annotation tool, we construct a large-scale dataset named as NWPU-Crowd~\cite{wang2020nwpu}, and more detailed description is shown in the video demo at \url{https://www.youtube.com/watch?v=U4Vc6bOPxm0/}. Moreover, some mainstream models are benchmarked on our NWPU-Crowd~\cite{wang2020nwpu} dataset, the code of which are open-sourced at \url{https://github.com/gjy3035/NWPU-Crowd-Sample-Code.} and more detailed results are at \url{ https://www.crowdbenchmark.com/nwpucrowd.html}.

As a whole, constructing cross-scene, multi-view and accurately annotated datasets which able to more faithfully reflect the real world challenge is essential to boost the generalization ability of crowd counting. Additionally, effective annotation tools are vital for the construction of the datasets.

\subsection{The quality of density maps}

Most existing methods pay attention to the count rather than the quality of density maps, which is an essential factor that affects the performance. Sindagi~\cite{sindagi2017generating} first observed this issue and proposed to incorporate global context into the training process while used adversarial loss together with Euclidean loss to obtain shaper and high-quality density maps. We choose several representing methods that concentrate the quality of density maps in terms of two measures, i.e., PSNR and SSIM~\cite{wang2004image}, as illustrated in Table~\ref{table:quality}. From the table, we can see that SE Cycle GAN~\cite{wang2019learning} shows the worst performance. We suspect it should be attributed to the ``domain gap'' between synthetic data and real-world data.

\begin{table}[!htb]

\caption{Comparison of the quality of density maps on the SHT\_A dataset~\cite{zhang2016single} in terms of PNSR and SSIM~\cite{wang2004image}. \mbox{\color{red}{\textbf{Red}}} and \underline{\textbf{underline}} indicate the best and the worst performance, respectively. "$\uparrow$" indicates the higher the better of the results.)}
\begin{center}
\begin{tabular}{|l|l|l|l|}
\hline
Methods & Year$\&$Venue & PSNR$\uparrow$ & SSIM$\uparrow$ \\ \hline
MCNN~\cite{zhang2016single} &2016 CVPR &21.4 &0.52 \\ \hline
Switching CNN~\cite{sam2017switching} &2017 CVPR &21.91 &0.67 \\ \hline
CP-CNN~\cite{sindagi2017generating} &2017 ICCV &21.72 &0.72 \\ \hline
CSRNet~\cite{li2018csrnet} &2018 CVPR &23.79 &0.76 \\ \hline
MPC~\cite{jiang2019effective} &2019 AI &21.24 &0.62 \\ \hline
SE Cycle GAN~\cite{wang2019learning} &2019 CVPR &\underline{\textbf{18.61}} &\underline{\textbf{0.407}} \\ \hline
DACC~\cite{gao2019domain} &2019 arxiv &21.94 &0.502 \\ \hline
W-Net~\cite{valloli2019w} &2019 arXiv &-- &\color{red}{\textbf{0.93}} \\ \hline
PCC Net~\cite{gao2019perspective} &2019 TCSVT &22.78 &0.74 \\ \hline
MVSAN~\cite{qiu2019crowd} &2019 ICME &23.17 &0.77\\ \hline
SCAR~\cite{gao2019scar} &2019 Neurocomputing &23.93 &0.81\\ \hline
CFF~\cite{shi2019counting} &2019 ICCV &25.4 &0.78 \\ \hline
ADCrowdnet~\cite{liu2019adcrowdnet} &2019 CVPR &24.48 &0.88 \\ \hline
TEDnet~\cite{jiang2019crowd} &2019 CVPR &\color{red}{\textbf{25.88}} &0.83 \\ \hline
DADNet~\cite{guo2019dadnet} &2019 ACMMM &24.16 &0.81 \\ \hline
ANF~\cite{zhang2019attentional} &2019 ICCV &24.1 &0.78 \\ \hline

\end{tabular}
\end{center}
\label{table:quality}
\end{table}

\subsection{Domain adaption or transfer learning}
Supervised learning requires accurate annotations, which is tedious by manually labeling, especially in extremely congested scenes. Also, mainstream models are almost designed for domain-specific. However, when generalizing the training model to unseen scenes, it would produce sub-optimal results due to the unpredictable domain gap. Table \ref{table:transfer} reports the evaluation results on Shanghai Part A \cite{zhang2016single} of the pre-trained NWPU-Crowd \cite{wang2020nwpu} models. Compared with the oracle errors, there are obvious performance degradations: the average MAE increases by \textcolor{red}{44.6\%} and RMSE increases by \textcolor{red}{47.0\%} respectively.

\begin{table}[!htb]
	\caption{The MAE/RMSE of the oracle evaluation (traditional supervised training within Shanghai Tech Part A) and cross-dataset evaluation (training on NWPU-Crowd and test on Shanghai Tech Part A), and the corresponding performance degradations.}
	\begin{center}
		\begin{tabular}{|c|c|c|c|c|c|}
			\hline
			Method & \makecell[c]{Oracle \\(MAE/RMSE)} & \makecell[c]{Cross-dataset \\Evaluation \\ (MAE/RMSE)} & \makecell[c]{Relative \\ Errors Rise (\%)} \\ \hline
			MCNN~\cite{zhang2016single} & 110.2/173.2 & 151.6/217.7 &$\uparrow$ \textcolor{red}{37.6/25.7}\\ \hline
			CSRNet~\cite{li2018csrnet} & 68.2/115.0 & 111.0/169.2 &$\uparrow$ \textcolor{red}{62.6/53.5}\\ \hline
			C3F-VGG~\cite{gao2019c} & 71.4/115.7 & 96.5/151.6 &$\uparrow$ \textcolor{red}{35.2/31.0}\\ \hline
			CANNet~\cite{liu2019context} & 62.3/100.0 & 83.5/137.4 &$\uparrow$ \textcolor{red}{34.0/37.4}\\ \hline
			SCAR~\cite{gao2019scar} & 66.3/114.1 & 96.6/161.5 &$\uparrow$ \textcolor{red}{45.7/71.5}\\ \hline
			SFCN\dag~\cite{wang2019learning} & 71.5/114.3 & 108.8/185.8 &$\uparrow$ \textcolor{red}{52.2/62.6}\\ \hline						
		\end{tabular}
	\end{center}
	\label{table:transfer}
\end{table}

The main reason about the performance degradations is that there are many differences (also named as domain gap/shifts) between the above datasets, including density range, image style, \emph{etc.} To remedy the domain gap, the technique of domain adaptation comes in handy, which can provide a feasible solution to reduce manpower by transferring effective features across diverse domains. In this process, GAN-based methods have demonstrated an important influence on this issue. For instance, SSIM Embedding (SE) Cycle GAN~\cite{wang2019learning} takes advantage of the domain adaptation technique, incorporating Structural Similarity Index (SSIM)~\cite{wang2004image} into traditional CycleGAN framework to make up the domain gap between synthetic data and real-world data. Although SE Cycle GAN~\cite{wang2019learning} addresses the problem to a certain extent, there still performs relatively low estimation count than other state-of-the-art methods. However, it paves the way for the transfer between different domains. We believe it will be of great benefit to crowd counting by refining this GAN-based method in the future.

\begin{figure}[t]
	\centering
	\includegraphics[width=0.95 \linewidth]{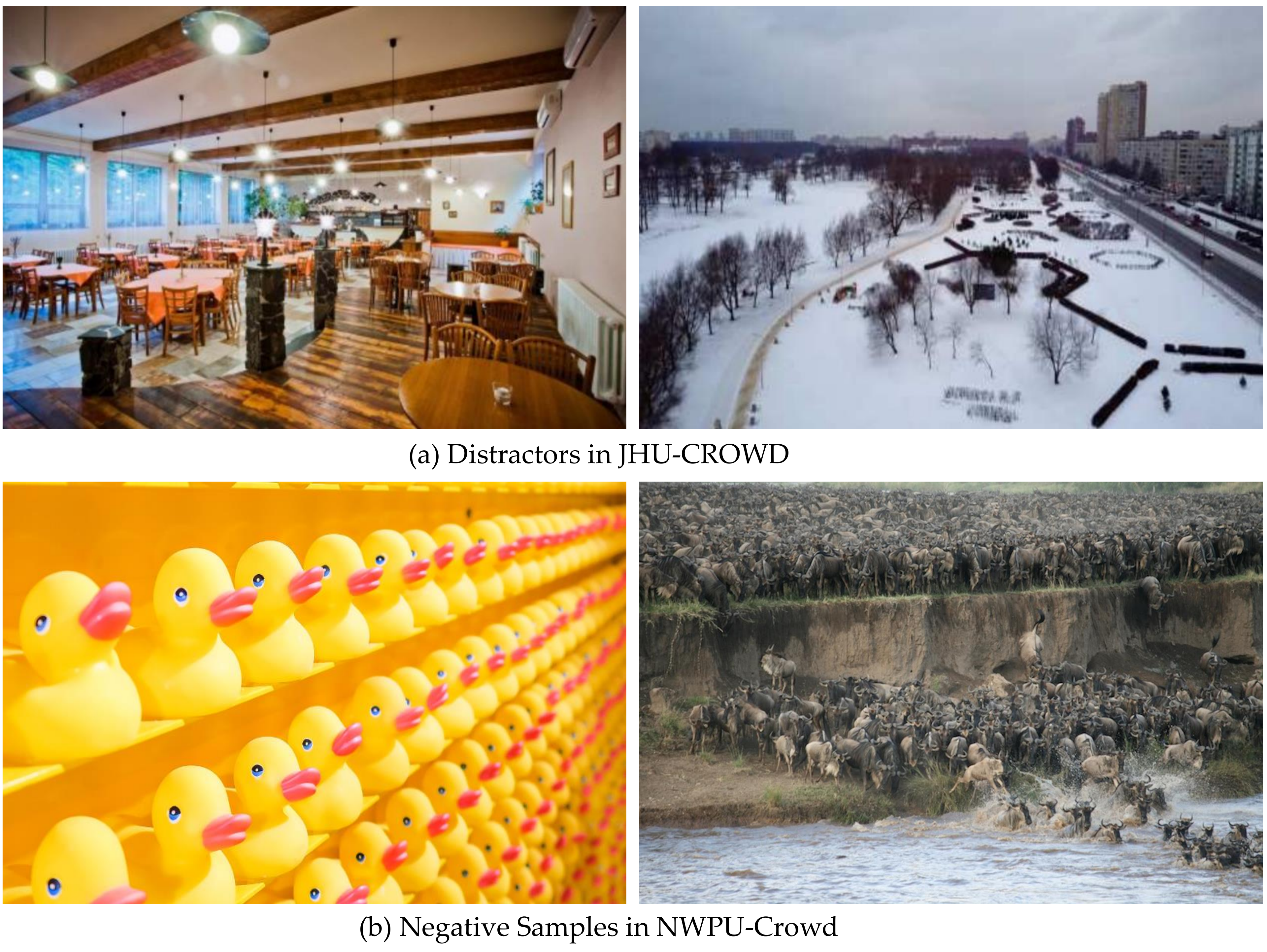}
	\caption{The exemplars of distractors and negative samples for crowd counting. }
	\vspace{-5pt}
	\label{fig:negative}
\end{figure}

\subsection{Robustness for background}

A robust counting model does not only accurately estimate the crowd density but also produce the zero-density response for background regions. To further evaluate models' robustness, the recently released large-scale datasets, such as JHU-CROWD \cite{sindagi2019pushing} introduces $100$ distractors and NWPU-Crowd \cite{wang2020nwpu} introduces $351$ negative samples into their own datasets, respectively.  These data do not contain person objects or crowd regions. It is worth mentioning that NWPU-Crowd \cite{wang2020nwpu} deliberately collects some scenes with densely arranged other objects to confuse counting models. Fig. \ref{fig:negative} shows some typical distractors and negative samples. These labeled data can effectively facilitate the counting model to perform better in the real world.

\begin{table}[!htb]
	\caption{The MAE/RMSE of mainstream methods on the distractors of JHU-CROWD and the negative samples of NWPU-Crowd.}
	\begin{center}
		\begin{tabular}{|c|c|c|c|c|}
			\hline
			Method & \makecell[c]{Distractors \\ in JHU-CROWD} & \makecell[c]{Negative Samples \\ in NWPU-Crowd} \\ \hline
			MCNN~\cite{zhang2016single} & 103.8/238.5 & 356.0/1232.5\\ \hline
			CMTL~\cite{zhang2016single} & 135.8/263.8 & -\\ \hline
			Switching CNN~\cite{sam2017switching} & 100.5/235.5 &-\\ \hline
			SA-Net~\cite{cao2018scale} & 71.9/167.7 & 432.0/974.4\\ \hline
			PCC-Net~\cite{gao2019perspective} & - & 85.3/438.8\\ \hline
			CSRNet~\cite{li2018csrnet} & 44.3/102.4 & 176.0/572.3\\ \hline
			C3F-VGG~\cite{gao2019c} & - & 141.0/474.2\\ \hline
			CANNet~\cite{liu2019context} & - & 82.6/343.4\\ \hline
			SCAR~\cite{gao2019scar} & - & 122.9/660.8\\ \hline
			SFCN\dag~\cite{wang2019learning} & - & \textbf{54.2/154.7}\\ \hline			
			CG-DRCN~\cite{sindagi2019pushing} & \textbf{43.4/97.8} &-\\ \hline

		\end{tabular}
	\end{center}
	\label{table:negative}
\end{table}

Table \ref{table:negative} lists the estimation errors (MAE/RMSE) on the aforementioned distractors and negative samples. From the results, we can find that current models mistakenly estimate the density of these samples. For a light model, PCC-Net performs better than many VGG-backbone methods (CSRNet~\cite{li2018csrnet}, C3F-VGG~\cite{gao2019c} and SCAR~\cite{gao2019scar}). The main reason may be reside in that PCC-Net~\cite{gao2019perspective} incorporates segmentation information to classify the foreground (namely head) and background. Therefore, there maybe an alternative way that uses multi-task learning (patch-level counting, segmentation, group detection, \emph{etc.}) to extract large-range features.

\subsection{Universality or generalization}
Nearly all the existing models for object counting are designed for a specific task, however, creating a universal model able to adapt any class of object is a meaningful challenge, it is also the most effective method to evaluate the robustness and generalization ability of the algorithm. Despite there are specialties between different tasks, there still exist many commonalities, such as crowd counting, vehicle counting, and cell counting. For instance, CAC~\cite{lu2018class} formulates the counting as a matching object through mining the self-similarity between images, which presents a Generic Matching Network (GMN) in a class-agnostic manner. PPPD~\cite{marsden2018people} provides a patch-based, multi-domain object counting network by leveraging a set of domain-specific scaling and normalization layers, which only uses a few of parameters. It can also be extended to perform a visual domain classification even in an unseen observed domain, which outstands out its versatility and modular nature. The method has been successfully applied to people, penguins, and cell counting.

Designing a unified principle (the generation of ground truth density map by using Gaussian function in ~\cite{lempitsky2010learning} takes a good example) and framework that can be applied to different tasks, it looks a bit awkward yet promising research direction in future.

\subsection{Lightweight network}

Current CNN-based deep models are designed with a sophisticated structure, which always comes with millions of parameters and the cost of a tremendous increase in computation (FLOPs). Although a great effort has been devoted to making the model efficient, such as CSRNet~\cite{li2018csrnet} and SCNet~\cite{wang2018defense}, they usually adopt VGG16~\cite{simonyan2014very} or ResNet~\cite{he2016deep} pre-trained on Imagenet~\cite{deng2009imagenet}, which is a large dataset for classification, but the task of object counting belongs to a regression task. Thus it may affect the performance to some extent. Additionally, the pre-trained mechanism is a very time-consuming process. Generally speaking, the most straightforward way to determine whether a network is lightweight is the number of parameters, the less the number of parameters, the lighter of the model. Table~\ref{table:parameters} shows a comparison of the number of parameters in some representative models. From the Table, we can find that LCNN~\cite{wu2019video} has the least number of parameters, which has nearly 2138 $\times$ lower than the worst one, CP-CNN~\cite{sindagi2017generating}. We suspect it should be attributed to LCNN~\cite{wu2019video} is a shallow network without pretraining. Less number of parameters, which proves more efficient of the model.

Lightweight networks can reduce the computation cost, but they are usually accompanied by accuracy drop. Therefore, in the premise of without sacrificing accuracy, designing lightweight and efficient networks to reduce the computation cost in the counting task is a promising challenge in the future.

\begin{table}[!htb]
\caption{Number of parameters (in millions). \mbox{\color{red}{\textbf{Red}}} and \underline{\textbf{underline}} indicate the least and the most parameters' number, respectively.}
\begin{center}
\begin{tabular}{|c|c|}
\hline
Method & Parameters \\ \hline
MCNN~\cite{zhang2016single} & 0.13 \\ \hline
Hydra-CNN~\cite{onoro2016towards} &0.56 \\ \hline
Switching CNN~\cite{sam2017switching} & 15.11 \\ \hline
CP-CNN~\cite{sindagi2017generating} & \underline{\textbf{68.4}} \\ \hline
CSRNet~\cite{li2018csrnet} & 16.26 \\ \hline
SA-Net~\cite{cao2018scale} & 0.91 \\ \hline
ACSCP~\cite{shen2018crowd} & 5.1 \\ \hline
L2R~\cite{liu2018leveraging} & 16.75 \\ \hline
DRSAN~\cite{liu2018crowd} &24.10 \\ \hline
ic-CNN~\cite{babu2018divide} &16.82 \\ \hline
IG-CNN~\cite{ranjan2018iterative} &4.70 \\ \hline
D-Conv1et~\cite{shi2018crowd} &16.62 \\ \hline
BSAD~\cite{huang2018body} &1.30 \\ \hline
MMCNN~\cite{yang2018counting} &6.8 \\ \hline
TDF-CNN~\cite{sam2018top} &1.15 \\ \hline
ASD~\cite{wu2019adaptive} &16.26 \\ \hline
TEDnet~\cite{jiang2019crowd} &1.63 \\ \hline
ANF~\cite{zhang2019attentional} &7.9 \\ \hline
MRCNet~\cite{bahmanyar2019mrcnet} &20.3 \\ \hline
SDA-MCNN~\cite{yang2019counting} &2.0 \\ \hline
LCNN~\cite{wu2019video} &\color{red}{\textbf{0.032}} \\ \hline
\end{tabular}
\end{center}
\label{table:parameters}
\end{table}

\subsection{Combination of image and video}
Modern mainstream models for counting have been deployed only for images or videos~\cite{zou2019enhanced,xiong2017spatiotemporal,zhang2017fcn,he2019dynamic,wu2019video,liu2019geometric,liu2019estimating}. When the video sequences are available, some algorithms are proposed to leverage temporal consistency to impel weak constraints on consecutive density estimation. Xiong et al.~\cite{xiong2017spatiotemporal} utilize an LSTM model to estimate densities from one frame to the next. Liu et al.~\cite{liu2019geometric} explicitly enforce the constraint under the condition that the number of people must be strictly conserved as they move about. Nevertheless, the constraint is difficult to express, w.r.t. densities. Furthermore, Liu et al.~\cite{liu2019estimating} regress people flow rather than regressing densities from video sequences, which imposes strong consistency constraints without complicated network architectures required. Therefore, designing effective algorithms, which can cater to images and videos simultaneous is a meaningful and promising direction.

\subsection{Wider-view crowd counting}
Albeit outstanding performance have been achieved for crowd counting in single-view images, it is not applicable to large and wide scenes such as public parks or long subway platforms, since it cannot capture sufficient detailed information for a single-view camera. Therefore, to address the problem of wide-area counting, some efforts have been attempted to capture information from multiple camera views. For instance, Zhang et al.~\cite{zhang2019wide} proposed a multi-view multi-scale (MVMS) fusion model to predict a 2D scene-level density map on the ground-plane. Furthermore, Zhang et al.~\cite{zhang20203d} achieve this by using 3D feature fusion with 3D scene-level density maps. Compared with 2D fusion~\cite{zhang2019wide}, 3D fusion not only preserves the property of 2D density maps but also extracts more useful information of the crowd densities along the z-dimension (height). The aforementioned two models are based on the assumption that the cameras are fixed and camera parameters are known, therefore, designing models for cross-scene and multi-view counting with moving cameras and unknown camera parameters is an interesting yet challenging future work.

\subsection{Localization, classification and tracking beyond object counting}
The density estimation CNN-based models for crowd counting, regression-based methods indeed, although accurate count provided, it does not indicate the precise location and exact size of objects, thus maybe limits the further research and application, such as high-level understanding, localization, classification, and tracking. Some attempts have been made, for instance, DecideNet~\cite{liu2018decidenet} generates the detection and regression-based density maps separately to estimate crowd density, and an attention module is incorporated to guide the final count. However, the model trains a fully supervised network with bounding box annotations, which needs to take large computation cost. CL~\cite{idrees2018composition} regresses density and localization maps simultaneously by introducing a composition loss. LCFCN~\cite{laradji2018blobs} estimates the crowd count by segmenting the object blobs in an image, which only employs the point-annotations. CL~\cite{idrees2018composition} and LCFCN~\cite{laradji2018blobs} simply concern the localization of crowds, whereas, PSDDN~\cite{liu2019point} not only predicts the localization but also estimates the size of persons. LSF-CNN~\cite{sam2019locate} locates the position of every person in the crowd, sizes the dot-annotated heads with bounding boxes and finally counts them. All the above works demonstrate that the potential research value of this direction.

\subsection{Small or tiny object counting}

The problem of small or tiny objects has long been a challenging task in many computer vision communities. In highly congested crowd scenes, the sizes of persons' heads are tiny. Additionally, some other potential applications may include object counting the number of contiguous dense buildings, ships, small vehicles and countless others in remote sensing images~\cite{gao2020dense}. An apparent difference between object counting in remote sensing scene and nature scenes, the orientations of the objects are arbitrary due to the overhead view rather than upright perspective. Some further directions may include the integration of visual attention mechanism, dilated convolution, deformable convolution layer, and rotation invariance design into the framework.

\section{Conclusion}
\label{section:conclusion}

Remarkable progress has been made in crowd counting over the past few decades. This paper has presented a survey of CNN-based density estimation and crowd counting models from several perspectives, including network architecture, learning paradigms, etc. We then summarize popular benchmark datasets, including crowd counting and several representing ones in other fields, as well as evaluation criteria for evaluating various methods. Besides, we also conduct a thorough performance benchmarking evaluation of representative models. Although all the works cannot be covered, we have selected the top-three performers to follow by a comprehensive and thorough analysis and discussion of these representing methods. We summarize the attributes or techniques which have a great assistant for the improvement of the performance.

In the next, we investigate several factors that would affect the performance of crowd counting, and we finally look through some potential challenges and open issues in deep learning era, and put forward insightful discussions and promising research directions in the future.

Standing on the standpoint of technical innovations, we expect this work can provide a feasible scheme to understand state-of-the-art, but more importantly, insights for future exploration in crowd counting and bridge to object counting in other domains.

\section*{Acknowledgment}

The authors would like to thank reviewers for their valuable suggestions and comments.

%-------------------------------------------------------------------------------------------------------------------------------------------------------------%
\bibliographystyle{IEEEtran}
\bibliography{IEEEabrv,references}

\begin{IEEEbiography}[]{Guangshuai Gao}
received the B.Sc. degree in applied physics from college of science and the M.Sc.
degree in signal and information processing from
the School of Electronic and Information Engineering, from the Zhongyuan University of Technology,
Zhengzhou, China, in 2014 and 2017, respectively.
He is currently pursuing the Ph.D. degree with
the Laboratory of Intelligent Recognition and Image
Processing, Beijing Key Laboratory of Digital Media, School of Computer Science and Engineering,
Beihang University. His research interests include
image processing, pattern recognition, and digital machine learning.
\end{IEEEbiography}
\begin{IEEEbiography}[]{Junyu Gao}
received the B.E. degree in computer
science and technology from the Northwestern Polytechnical University, Xi’an 710072, Shaanxi, P. R.
China, in 2015. He is currently pursuing the Ph.D.
degree from Center for Optical Imagery Analysis
and Learning, Northwestern Polytechnical University, Xian, China. His research interests include
computer vision and pattern recognition.
\end{IEEEbiography}
\begin{IEEEbiography}[]{Qingjie Liu}
  received the Ph.D. degree in computer
science from the Intelligent Recognition and Image
Processing Laboratory, Beihang University, Beijing,
China, in 2015.
He is currently an Assistant Professor with the
School of Computer Science and Engineering, Beihang University. He is also a Distinguished Research
Fellow with the Hangzhou Institute of Innovation,
Beihang University, Hangzhou. His current research
interests include Remote sensing image analysis,
pattern recognition, and computer vision. He is a
member of the IEEE.
\end{IEEEbiography}

\begin{IEEEbiography}[]{Qi Wang}
(M’15-SM’15) received the B.E. degree in
automation and the Ph.D. degree in pattern recognition and intelligent systems from the University
of Science and Technology of China, Hefei, China,
in 2005 and 2010, respectively. He is currently a
Professor with the School of Computer Science and
with the Center for OPTical IMagery Analysis and
Learning (OPTIMAL), Northwestern Polytechnical
University, Xi’an, China. His research interests include computer vision and pattern recognition.
\end{IEEEbiography}

\begin{IEEEbiography}[]{Yunhong Wang}
received the B.S. degree in electronic engineering from Northwestern Polytechnical
University, Xi’an, China, in 1989, and the M.S.
and Ph.D. degrees in electronic engineering from
the Nanjing University of Science and Technology,
Nanjing, China, in 1995 and 1998, respectively.
She was with the National Laboratory of Pattern Recognition, Institute of Automation, Chinese
Academy of Sciences, Beijing, China, from 1998 to
2004. Since 2004, she has been a Professor with
the School of Computer Science and Engineering,
Beihang University, Beijing, where she is also the Director of the Laboratory
of Intelligent Recognition and Image Processing and the Beijing Key Laboratory of Digital Media. Her research interests include biometrics, pattern
recognition, computer vision, data fusion, and image processing. She is a
Fellow of IEEE, IAPR, and CCF.
\end{IEEEbiography}

\end{document}